\pdfoutput=1
\documentclass[11pt]{article}
\usepackage{acl}

\usepackage{times}
\usepackage{latexsym}
\usepackage[T1]{fontenc}
\usepackage[utf8]{inputenc}
\usepackage{microtype}
\usepackage{inconsolata}
\usepackage{arydshln}
\usepackage{graphicx}
\usepackage{tabularx,booktabs}

\usepackage{graphicx}
\usepackage{hyperref}
\usepackage{multirow}
\usepackage{graphicx}
\usepackage{tabularx}
\usepackage{url}
\usepackage{xcolor}
\usepackage{epsfig}
\usepackage{adjustbox}
\usepackage{amsfonts, amsmath, amssymb}
\usepackage{booktabs} 
\usepackage{comment}
\usepackage{caption, subcaption}
\usepackage{textcomp}
\usepackage{relsize}
\usepackage{stmaryrd}
\usepackage{bbm}
\usepackage{rotating}
\usepackage{helvet}
\usepackage{courier}
\usepackage{cleveref}
\usepackage{xspace}
\usepackage{enumitem}
\usepackage{epigraph}
\usepackage{mdframed}
\usepackage{makecell}
\usepackage{tcolorbox}
\usepackage{nicematrix}
\usepackage{tabu}
\usepackage{colortbl}
\usepackage{xcolor}
\PassOptionsToPackage{table}{xcolor}

\usepackage{pifont}

\usepackage{listings}
\newcolumntype{X}{>{\raggedright\arraybackslash}m{0.9\linewidth}}
\newcolumntype{W}{>{\centering\arraybackslash}m{0.06\linewidth}}
\newcolumntype{L}{>{\arraybackslash}m{0.99\linewidth}}

\newcommand{\task}{\textsc{MT-RAIG}\xspace}
\newcommand{\fulltask}{Retrieval-Augmented Insight Generation over Multiple Tables}

\newcommand{\bench}{\textsc{MT-RAIG Bench}\xspace}

\newcommand{\wikidataset}{Open-WikiTable\xspace}

\newcommand{\eval}{\textsc{MT-RAIG Eval}\xspace}

\title{\includegraphics[width=0.8em]{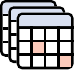}\task: Novel Benchmark and Evaluation Framework for \fulltask}

\author{
    Kwangwook Seo\thanks{\; Equal contribution}~~~
    Donguk Kwon$^\ast$~~~
    Dongha Lee\thanks{\; Corresponding author}\\
    Yonsei University - DLI Lab\\
    \texttt{\{tommy2130,donguk.kwon,donalee\}@yonsei.ac.kr}\\   
}

\begin{document}
\maketitle
\begin{abstract}
Recent advancements in table-based reasoning have expanded beyond factoid-level QA to address insight-level tasks, where systems should synthesize implicit knowledge in the table to provide explainable analyses. Although effective, existing studies remain confined to scenarios where a single gold table is given alongside the user query, failing to address cases where users seek comprehensive insights from multiple unknown tables. To bridge these gaps, we propose \textbf{\textsc{MT-RAIG Bench}}, designed to evaluate systems on \textbf{R}etrieval-\textbf{A}ugmented \textbf{I}nsight \textbf{G}eneration over \textbf{M}ultiple-\textbf{T}ables.
Additionally, to tackle the suboptimality of existing automatic evaluation methods in the table domain, we further introduce a fine-grained evaluation framework \textbf{\eval}, which achieves better alignment with human quality judgments on the generated insights. 
We conduct extensive experiments and 
reveal that even frontier LLMs still struggle with complex multi-table reasoning, establishing our \bench as a challenging testbed for future research\footnote{\url{ https://kwondu.github.io/mt-raig}}.
\end{abstract}

\section{Introduction}
\label{sec:introduction}
\begin{figure}[ht]
\centering
\includegraphics[width=\linewidth]{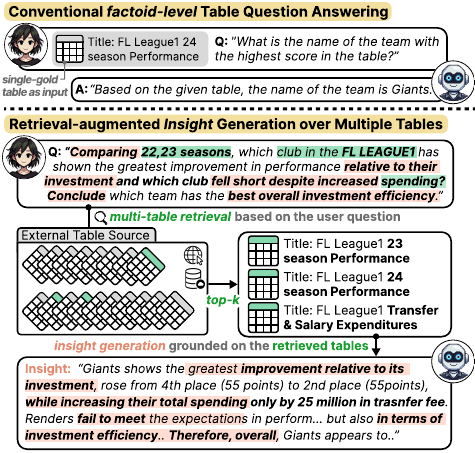}
\caption{Comparison between \task and conventional table question answering task. }
\vspace{-0.3cm}
\label{fig:motivation}
\vspace{-0.35cm}
\end{figure}

Tables are ubiquitous in the real-world data and constitute a significant portion of the information available on the web and databases. 
While their structured nature efficiently encapsulates diverse information, it also poses challenges for developing robust table understanding systems~\citep{pasupat-liang-2015-compositional,chen-etal-2020-hybridqa,tang2024strucbenchlargelanguagemodels,Seo2024UnveilingIT}.
This inherent complexity of table data has led to a persistent demand for the systems capable of faithfully interpreting table content and presenting it to users in a human-readable format.

In response to these needs, existing table-based question answering (TQA) works~\citep{pasupat-liang-2015-compositional,Nan2021FeTaQAFT,Wang2024ChainofTableET} have predominantly focused on extracting explicit facts from a given table by developing systems that can follow the detailed instructions outlined in user queries.
However, these \textit{factoid-level} queries often constrain the systems’ operation to functioning as an extractive executor, reducing its role to retrieving only a fraction of values explicitly presented in the table. 
This narrow focus prevents the systems from comprehensively analyzing the table’s full context, which is essential for uncovering  implicit information embedded within the table.

Recently, some studies~\citep{moosavi2021scigen,Zhao2023QTSummAN,Seo2024UnveilingIT} have begun to explore more practical user information needs by moving beyond this traditional \textit{factoid-level} scenario. 
These studies address \textit{insight-level} table reasoning, wherein the user requires not just simple fact retrieval but also deeper insight mining from the table. 
This shift has led to the development of advanced agents designed to deliver explainable analyses and meaningful data insights, accommodating scenarios where users seek comprehensive interpretations and synthesized knowledge.

\definecolor{darkgreen}{HTML}{006400}
\begin{table*}[!ht]
\centering
\begin{small}
\begin{tabularx}{\textwidth}{lcccccc}
\toprule
\textbf{Dataset} & 
\makecell[c]{\#\textbf{Test}\\ \textbf{Examples}} & 
\makecell[c]{\#\textbf{Unique}\\ \textbf{Tables}} & 
\makecell[c]{\textbf{\#Words}\\ \textbf{/Output}} &
\makecell[c]{\textbf{\#Tables}\\ \textbf{/Example}} & 
\makecell[c]{\textbf{Retrieval}\\ \textbf{Availability}} &
\makecell[c]{\textbf{Reasoning}\\ \textbf{Depth}} \\
\midrule
\multicolumn{7}{l}{\textit{Text-to-SQL generation}} \\
WikiSQL~\citep{zhong2018seqsql}
& 15,878 & 26,531 & 6.13 & 1.00
& \color{red} \ding{55} & \textit{factoid-level} \\
SPIDER~\citep{Yu2018SpiderAL}
& 1,034 & 876 & 18.37 & 1.58
& \color{red} \ding{55} & \textit{factoid-level} \\
\midrule
\multicolumn{7}{l}{\textit{Table-to-text generation}} \\
ToTTo~\citep{Parikh2020ToTToAC}
& 7,700 & 83,141 & 17.37 & 1.00
& \color{red} \ding{55} & \textit{factoid-level} \\
InsTaSumm~\citep{Seo2024UnveilingIT}
& 440 & 2,494 & 161.90 & 1.00
& \color{red} \ding{55} & \textit{insight-level} \\
\midrule
\multicolumn{7}{l}{\textit{Table question-answering}} \\
WTQ~\citep{Pasupat2015CompositionalSP}
& 9,473 & 2,108 & 1.76 & 1.00
& \color{red} \ding{55} & \textit{factoid-level} \\
FeTaQA~\citep{Nan2021FeTaQAFT}
& 2,003 & 10,330 & 23.30 & 1.00
& \color{red} \ding{55} & \textit{factoid-level} \\
Open-WikiTable~\citep{Kweon2023OpenWikiTableDF}
& 6,602 & 24,680 & 1.90 & 1.00
& \color{darkgreen} \ding{51} & \textit{factoid-level} \\
QTSumm~\citep{Zhao2023QTSummAN}
& 1,078 & 2,934 & 67.76 & 1.00
& \color{red} \ding{55} & \textit{insight-level} \\
\midrule
\textbf{\bench} (ours)
& \textbf{18,532} & \textbf{19,563} & \textbf{189.87} & \textbf{2.88} 
& \color{darkgreen} \ding{51} & \textit{\textbf{insight-level}} \\
\bottomrule
\end{tabularx}
\end{small}
\caption{Comparison with existing benchmarks for table-related tasks.}
\label{tab:comparison}
\end{table*}

Despite these advancements in table-based reasoning, two critical limitations remain unaddressed. 
\textbf{(1)} Existing approaches typically operate in a \textbf{closed-domain setting}, where a pre-defined gold table is provided alongside the query at test time. 
While it simplifies the testbed for evaluating the system's reasoning ability, it also introduces a significant drawback. 
In particular, requiring users to manually craft table inputs for every query is both costly and unrealistic, as users often lack prior knowledge about which specific tables are relevant to their needs.
\textbf{(2)} Most prior works focus on \textbf{single-table tasks}, assuming all source information for the reasoning is contained within a single table.
However, considering that users' information needs may require comprehensive insights spanning multiple aspects across tables, this scenario falls short in robustly handling diverse user needs.

To bridge these gaps, we propose \textbf{\textsc{MT-RAIG Bench}}, aiming to measure the system’s ability on \textbf{R}etrieval-\textbf{A}ugmented \textbf{I}nsight \textbf{G}eneration over \textbf{M}ulitple-\textbf{T}ables.
To enable more practical applications of the table-based system, our \textsc{MT-RAIG Bench} requires the system to retrieve multiple evidence tables based on the input query and integrate information across them to generate insightful response.
Compared to existing benchmarks, this task introduces new challenges by requiring the systems not only to retrieve query-relevant tables but also to faithfully extract evidences scattered among irrelevant tables and comprehensively aggregate these evidences to derive meaningful insight. 

Although \textsc{MT-RAIG Bench} can serve as a promising testbed for insight-level table reasoning, reliably evaluating long-form outputs in table-based tasks remains a longstanding challenge in the field~\citep{zhao-etal-2024-tapera}.
While recent studies~\citep{zhao-etal-2023-investigating,wang-etal-2024-revisiting} employ table-specific metrics that assess the quality of generated outputs beyond the surface-level matching, they still struggle to evaluate the quality of the insight on the MT-RAIG task.
Such limitation arise from their reliance on coarse-grained analyses, which fall short in detecting the finer distinctions needed to check both the output’s groundness on multi-tables and completeness on a multi-hop query. 
\begin{figure*}[!ht]
\centering
\includegraphics[width=1\textwidth]{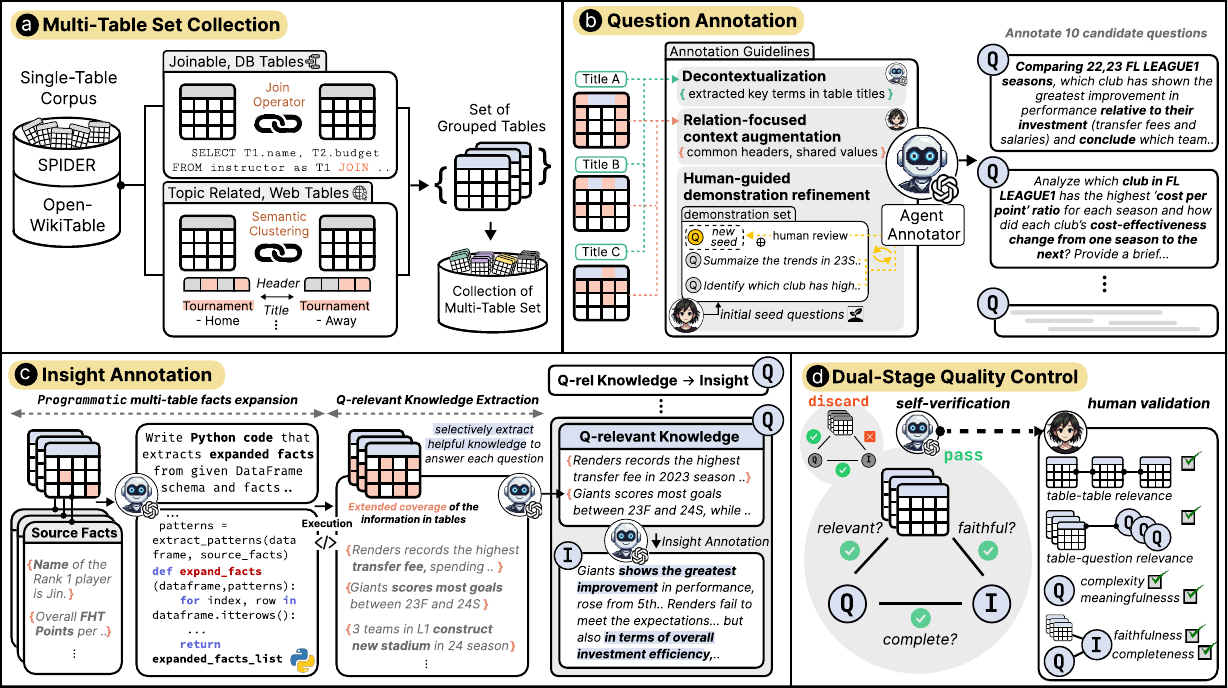}
\caption{Overview of our \bench construction process.}
\label{fig:benchmark}

\end{figure*}

In light of these challenges, we propose a novel decomposition-based evaluation framework \textbf{\textsc{MT-RAIG Eval}}.
To reliably assess the finer quality of long-form insights in \textsc{MT-RAIG Bench}, \textsc{MT-RAIG Eval} performs: (1) \textit{table-aware insight decomposition} to verify the explicit grounding between the fact-entangled insight and retrieved tables, (2) \textit{question-aware insight decomposition} to check whether the key steps required to address the question are completely followed in the insight. 
Our meta-evaluation validates that \textsc{MT-RAIG Eval} outperforms conventional metrics in aligning with human judgments, reliably assessing both faithfulness and completeness of generated insights.
We summarize our contributions as follows:
\begin{itemize}[leftmargin=*,topsep=2pt,itemsep=2pt,parsep=0pt]
    \item We propose \bench, the first large-scale benchmark for retrieval-augmented insight generation over multiple tables.
    \item We introduce \textsc{MT-RAIG Eval}, a novel automated evaluation framework that assesses the fine-grained quality of the generated insights.
    \item We evaluate various LLMs and SOTA methods, revealing that existing models still struggle with multi-table reasoning, establishing our benchmark as a challenging testbed for future research.
\end{itemize}

\section{\bench}
\label{sec:benchmark}
\subsection{Task Formulation}
In line with the recent advancements of retrieval-augmented generation (RAG) in diverse domains~\citep{ram-etal-2023-context,shi-etal-2024-replug}, \textsc{MT-RAIG} task consists of two main steps: (1) \textit{table retrieval}, (2) \textit{insight generation}. 
Formally, given a natural language question $q$ and an external datasource with tables $T$, the task is to first retrieve a set of evidence tables $\hat{T_{q}} \subset T$ to approximate the gold table set $T_{q}$ relevant to $q$. Subsequently, an insight $i$ is generated by grounding it on the retrieved $\hat{T_{q}}$. These two steps can be formulated as follows:
\begin{equation}
    \hat{T_q} = Ret(q, T), \quad i = Gen(q, \hat{T_q})
\end{equation}

\subsection{Benchmark Construction}
We provide detailed information on the construction of the \bench.
Specifically, we describe the process of creating a machine-annotated dataset, building on efforts from previous works~\citep{tang2024multihoprag,wei2024longform,ni2024mixeval,kim2024evaluatinglanguagemodelssynthetic,yao2024taubenchbenchmarktoolagentuserinteraction} while leveraging several notable advantages over fully human-annotated benchmark. 
These benefits include: \textbf{(1) \textit{scalability}}, where large-scale expansion is possible with significantly reduced human labor, 
and \textbf{(2) \textit{consistency}} in labeling standards, a crucial aspect that can be more challenging to maintain when multiple human annotators are involved.
Additionally, we incorporate human quality checks as critical review points for the automatically generated data, balancing the efficiency of automated annotation with the reliability of human annotation to ensure the high-quality of \bench.
We provide all the detailed prompts in Appendix~\ref{apx:annotate_detail} and benchmark examples in Appendix~\ref{apx:case_study}.

\subsubsection{Multi-Table Set Collection}
The goal of this step is to extend the existing single-table corpus into a collection of multi-table sets. 
To group individual tables in the data source into coherent sets that are relevant to real user queries, we classify scenarios where multiple tables are combined into two categories: \textit{joinable} and \textit{topic-related}.
Joinable tables can be directly linked through common key columns, following similar concept to standard text-to-SQL tasks. 
To construct such sets, we use SPIDER~\cite{Yu2018SpiderAL} as a source dataset, leveraging foreign key connections to identify joinable tables from single-table corpus. 

While joinable tables offer a straightforward approach to combining individual tables, real-world table sets often present more complex relationships, where tables cannot be directly joined via a common key column. Instead, these tables are loosely connected by shared topics or contextual relevance.
To effectively group these tables, we leverage table titles and headers as semantic indicators to cluster topically related tables.  
We source these tables from \wikidataset~\cite{Kweon2023OpenWikiTableDF}, a collection of tables extracted from Wikipedia, where rich metadata provides the necessary cues for collecting topically coherent multi-table sets.

\subsubsection{Question Annotation} 
Unlike existing factoid-level questions, our approach aims to annotate  questions that seek comprehensive insights across tables. 
Following previous work~\citep{kim2024evaluatinglanguagemodelssynthetic}, we employ GPT-4o mini as an agent annotator, and generate 10 distinct questions for each table set, designed to capture diverse relational aspects among the tables. 
To strike a robust middle ground between the reliability of human annotation and the versatility of LLMs, we adopt a \textit{human-in-the-loop} process that faithfully guides the question annotation while preserving the flexibility of LLMs. 
Specifically, the question annotation process incorporates three key methods:

\textbf{(1) Decontextualization}:
The goal of decontextualization~\citep{ Choi2021DecontextualizationMS,Kweon2023OpenWikiTableDF} is to enhance the clarity of how each question links back to the relevant tables. 
By explicitly including keywords derived from the table titles into the questions, this step ensures that the semantic alignment between the question and the tables become apparent. 
To achieve this, the agent annotator first extracts key terms from table titles and incorporate them directly into the phrasing of the questions. 

\textbf{(2) Relation-focused context augmentation:}
This step aims to highlight relationships between tables by enriching the questions with contextual information shared across the tables.
We find that naively prompting the agent annotator to generate questions for multiple sources often leads to mere concatenations of information separately extracted from each table, resulting in questions that fail to reflect the connections between the tables.
To address this issue, we first manually identify common attributes and shared data points across multiple tables and let the agent annotator incorporate these overlapping values into the question generation to ensure that critical relational cues between the tables are preserved within the generated questions.

\textbf{(3) Human-guided  demonstration refinement:}
To guide the agent in generating questions that are both diverse and aligned with our annotation objectives, we adopt an iterative workflow that combines agent-based automation with human feedback. 
We begin by categorizing questions into distinct types as shown in Table~\ref{tab:statistics}, and annotate initial human-written questions for each type.
These seed questions serve as initial demonstrations for agent annotator to generate new questions. Then, if human reviews confirm a novel pattern from generated questions, it is incorporated back into the seed set.
Through this iterative human-guided refinement, we build a final demonstration set to help the agent generate well-grounded questions for each type while covering diverse multi-hop relationships.

\begin{table}[h]
\centering
\begin{small}
\begin{tabularx}{\linewidth}{p{0.6\linewidth}>{\raggedleft\arraybackslash}X}
\toprule
\textbf{Properties}          & \textbf{Value}  \\ \midrule
\# of Test examples               & 18,532          \\
\# of Words / insight  & 189.87          \\ 
\# of Unique tables                & 19,563          \\
\# of Unique table set             & 5,418           \\
\# of Gold tables / example    & 2.88            \\
\# of Rows / table          & 10.54           \\
\# of Columns / table       & 6.04            \\ \midrule
\textbf{Question Types}      & \textbf{Size}   \\ \midrule
Analysis \& Summary (A\&S)          & 1,886 (10\%)    \\
Comparison \& Relationship (C\&R)   & 4,035 (22\%)    \\
Performance \& Outcome (P\&O)     & 10,141 (55\%)   \\
Trend \& Pattern (T\&P)        & 2,470 (13\%)    \\
\bottomrule
\end{tabularx}
\end{small}
\vspace{-0.2cm}
\caption{Basic statistics of \bench}
\label{tab:statistics}
\vspace{-0.3cm}
\end{table}

\subsubsection{Insight Annotation}
\paragraph{Programmatic multi-table facts expansion}
To enhance the agent annotators' understanding of structured data, we augment multi-table set with natural language (NL) facts that provide  additional context for each table set.
These facts are initialized from human-annotated sources in existing datasets (SPIDER, Open-WikiTable), which are accurate but focus narrowly on specific table portions, lacking comprehensive coverage of the full table context. 
To address this, we employ a programmatic fact expansion process to cover a broader range of information within the tables. 
Specifically, given the source tables and initial NL facts, the agent annotator generates a Python function, \texttt{expand\_facts}. 
This function is then executed on the tables to systematically extract enriched facts while ensuring faithfulness to the source table.

\paragraph{Question-relevant knowledge extraction}
Based on the enriched facts obtained in the previous step, the next step involves annotating insights that serve as a comprehensive answer to the given question.  
However, even when providing the agent annotator with enriched NL contexts derived from fact expansion, such contexts might introduce noise and hinder the insight annotation, as not all information helps address each question.
Therefore, we filter out irrelevant content for each question and selectively extract only the relevant facts to provide the agent annotator with a condensed form of knowledge.
This knowledge is then fed to the agent annotators to generate insights for each table set.

\subsubsection{Dual-Stage Quality Control}
For the last step of our benchmark construction, we adopt dual-stage quality assurance process that combines human and agent verification. 
First, following ~\citet{tang2024multihoprag}, we utilize an agent annotator as a self-verifier, ensuring that each multi-table set, question, and insight triple satisfies strict criteria for \textit{relevance}, \textit{faithfulness}, and \textit{completeness}—discarding any triple that fails at least one of these standards.
Second, we conduct a human validation, wherein the machine-verified samples are manually reviewed according to the criteria in Table~\ref{tab:validation}, thereby confirming the accuracy and coherence of the agent-based annotations. 
Table~\ref{tab:validation} summarizes this result, showing high degree of agreement among human evaluators, demonstrating the high-quality of our benchmark.
We provide the case study, statistics of self-verification, and detailed process of quality control in Appendix~\ref{apx:verify_detail}.

\begin{table}[h]
\centering
\begin{small}
\begin{tabularx}{\linewidth}{%
l
>{\centering\arraybackslash}X
>{\centering\arraybackslash}X
>{\centering\arraybackslash}X}
\toprule
\textbf{Data Quality} & \textbf{\%S$\geq$4} & \textbf{Agree} & \textbf{Kappa} \\ \midrule
Inter-Table Relevance & 97.33 & 0.83 & 0.78 \\
\noalign{\vskip 0.5ex}
\cdashline{1-4} 
\noalign{\vskip 0.5ex}
Table-Question Relevance & 96.00 & 0.87 & 0.84 \\
Question Complexity & 98.67 & 0.87 & 0.83 \\
Question Meaningfulness & 98.00 & 0.89 & 0.86 \\
\noalign{\vskip 0.5ex}
\cdashline{1-4}
\noalign{\vskip 0.5ex}
Question-Insight Completeness & 94.00 & 0.85 & 0.81 \\
Table-Insight Faithfulness & 89.67 & 0.86 & 0.82 \\
\bottomrule
\end{tabularx}
\end{small}
\vspace{-0.2cm}
\caption{Human evaluation over 300 samples of \bench. Two evaluators are asked to rate the samples in 1-5 Likert Scale. We report percentage of samples that have average score above 4, percent of evaluator agreement and Cohen’s Kappa with 95\% CI.}
\label{tab:validation}
\vspace{-0.5cm}
\end{table}

\section{\eval}
\label{sec:evaluation}
Automatic evaluation on the quality of long-form output in table-based tasks remains a long-standing challenge that often does not align well with human evaluations~\citep{zhao-etal-2024-tapera,wang-etal-2024-revisiting,Seo2024UnveilingIT}.
This issue is even more pronounced in the MT-RAIG task, as it is challenging to check (1) the explicit grounding between fact-entangled insight and multiple tables, (2) whether the key steps required to address the multi-hop questions are completely followed in insight.
We posit that these challenges arise as existing automatic evaluation methods (\textit{e.g.,} BLEU~\citep{Papineni2002BleuAM}, TAPAS-Acc~\citep{Liu2022PLOGTP}, and G-Eval~\citep{Liu2023GEvalNE}) often analyze the output in a coarse-grained manner. 
To tackle this challenge, we propose a novel \textbf{decomposition-based evaluation framework} \textbf{\textsc{MT-RAIG Eval}} that enables finer distinctions in assessing the quality of the long-form outputs. 
We focus on evaluating the following two key dimensions:

\begin{figure}[ht]
\centering
\includegraphics[width=\linewidth]{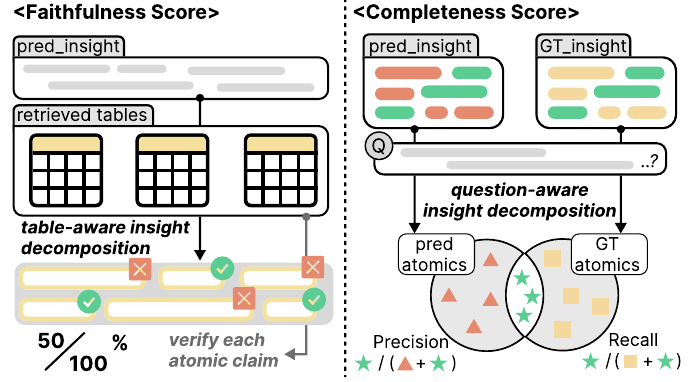}
\vspace{-0.2cm}
\caption{Overview of \eval}
\label{fig:evalutation}
\vspace{-0.3cm}
\end{figure}
\paragraph{Faithfulness Score} 
Faithfulness score evaluates whether an insight is fully grounded in the provided source tables.
Ideally, a perfect evaluator should assess whether all atomic facts entangled within an insight are correctly grounded in the retrieved tables.
To systematically verify these atomic facts, we introduce \textbf{\textit{table-aware insight decomposition}}, which breaks down an insight into verifiable claims. 
Specifically, we leverage an LLM-based decomposer, enhanced with structural guidance from table schemas, to generate a set of granular claims explicitly linked to their originating tables.
Each claim is then validated by an LLM verifier against the retrieved tables.
This table-aware decomposition enables a fine-grained evaluation of the predicted insights by ensuring traceability and reducing ambiguity of the verifiable claims in multi-table contexts.
Formally, given a set of retrieved tables $\hat{T_q}$, a predicted insight $i$,  a decomposer $\mathcal{D}$, and a verifier $\mathcal{V}$, the final score $S_{Faith.}$ is computed as follows:
\begin{equation}
    C = \mathcal{D}(\hat{T_q}, i), \ S_{Faith.}= \frac{1}{|C|} \sum_{k=1}^{|C|} \mathcal{V}(c_k, \hat{T_q})
\end{equation}
where $C = \{c_k\}_{k=1}^{n}$ is a set of decomposed claims, and $\mathcal{V}(c_k, \hat{T_q}) \in \{0, 1\}$ verifies each claim against the retrieved tables $\hat{T_q}$.

\paragraph{Completeness Score}
Completeness refers to the idea that a generated insight should fully address all the requirements outlined in the given question.
Independent of faithfulness, this dimension penalizes outputs that omit key analytical steps or introduce redundant content that deviates from the question’s intent.
To verify that these key steps are properly addressed, we propose a \textbf{\textit{question-aware insight decomposition}}, which deconstructs both the ground truth insight $i$ and the predicted insight $\hat{i}$ into atomic topics representing key steps necessary to resolve the question (\textit{e.g.,} identifying causal relationships or synthesizing cross-table comparisons). 
Specifically, we first employ an LLM-based decomposer $\mathcal{D}$, conditioned on the input question $q$, to generate two sets of atomic topics $A = \mathcal{D}(q, i)$ and $\hat{A} = \mathcal{D}(q, \hat{i})$. 
Subsequently, we perform a semantic matching by using an LLM $\mathcal{M}$, between the atomic topics in $A$ and $\hat{A}$ to compute the precision $P$ and recall $R$ based on the degree of overlap between these sets.   
The final completeness score $S_{Comp.}$ is the F1 score, formulated as follows:
\begin{equation}
\small 
    P = \frac{|\mathcal{M}(A, \hat{A})|}{|\hat{A}|}, \   
    R = \frac{|\mathcal{M}(A, \hat{A})|}{|A|}, \   
    F1 = \frac{2 \cdot P \cdot R}{P + R} 
\end{equation}

\section{Experiments}
\label{sec:experiments}
\subsection{Baselines}
We benchmark the performance of diverse baselines on \textsc{MT-RAIG Bench}.
For multi-table retrieval, we consider both \textit{general-purpose} retriever ~\citep{Robertson2009ThePR,Karpukhin2020DensePR,Izacard2021UnsupervisedDI}, and \textit{table-specific} retriever~\citep{Herzig2021OpenDQ,Zhang2023TableLlamaTO} as the baselines. 
For insight generation, we employ 11 LLM-based baselines including \textit{proprietary}~\citep{O3mini,GPT4o,Claude3S}, \textit{open-source}~\citep{DeepSeekAI2025DeepSeekR1IR,Yang2024Qwen2TR,Mesnard2024GemmaOM,llama-3,Jiang2023Mistral7}, and \textit{SOTA-TQA} methods~\citep{Ye2023LargeLM,Wang2024ChainofTableET,zhao-etal-2024-tapera} for comprehensive evaluations.
The results are shown in Table~\ref{tab:retriever} and Table~\ref{tab:generator}, respectively.
Please refer to Appendix~\ref{apx:experiments} for the detailed information on the experimental setup and all the baselines.

\subsection{Results}
\paragraph{Multi-table Retrieval Evaluation} To understand the challenges of multi-table retrieval in \bench, we compare the performance of five widely-used retrievers. 
From the results in Table~\ref{tab:retriever}, we observe that there remains a significant gap between high and low top‑$k$ performance,  which is critical for the generation performance given that LLMs often hold only a limited number of tables in their context window.
We also find that table-specific embedding models do not improve retrieval performance on \bench. Instead, general text-based embedding model achieves better performance, similar to the findings of \citet{wang-etal-2022-table}. 
We attribute this to the nature of the MT-RAIG task, which prioritizes identifying insight-level semantic connection of tables for multi-hop queries rather than understanding the structural characteristics of tables.
Based on these findings, we use DPR’s top-10 retrieved tables (considering the LLM’s context window limitation) for experiments in Table~\ref{tab:meta_main} and Table~\ref{tab:generator}.

\begin{table}[htb]
\centering
\small
\begin{tabularx}{\linewidth}{l l *{4}{>{\centering\arraybackslash}X}}
\toprule
\textbf{Type} & \textbf{Retriever} & \textbf{R@2} & \textbf{R@5} & \textbf{R@10} & \textbf{R@20} \\
\midrule
\multirow{3}{*}{\shortstack{\textit{general-}\\\textit{purpose}}}
 & BM25       & 17.26 & 27.19 & 33.72 & 41.16 \\
 & DPR        & \textbf{44.58} & \textbf{68.24} & \textbf{80.83} & \textbf{88.45} \\
 & Contriever & 23.47 & 35.67 & 44.27 & 52.12 \\
\midrule
\multirow{2}{*}{\shortstack{\textit{table-}\\\textit{specific}}}
 & DTR        & \underline{37.77} & \underline{59.60} & \underline{74.50} & \underline{86.22} \\
 & TableLlama & 36.93 & 59.48 & 72.44 & 81.56 \\
\bottomrule
\end{tabularx}
\vspace{-0.2cm}
\caption{Multi-table retrieval results on \bench. We report Recall at various top-\(k\). Full results including Precision and F1 are in Appendix~\ref{apx:results}.}
\label{tab:retriever}
\vspace{-0.3cm}
\end{table}

\paragraph{Meta Evaluation}
We conduct a meta evaluation to assess the reliability of \textsc{MT-RAIG Eval} against existing automatic metrics.
Specifically, we construct a meta evaluation dataset comprising 250 pairs of responses sampled from baseline generators, where each pair is labeled by two human evaluators based on their relative preferences on response faithfulness and completeness.
For scoring, we follow the setting of \citet{ru2024ragchecker} to normalize each auto-evaluation score difference to the human preference scale of [-1,0,1] and then measure the Pearson correlation with human preference ratings. 
Additionally, we report the correlation between the human evaluators as the upper bound. 
From the results in Table~\ref{tab:meta_main}, we can observe that \eval achieves the highest correlation with human preference ratings across both dimensions, demonstrating its reliability over baseline methods for evaluating \textsc{MT-RAIG Bench}. 
We provide more detailed results in Appendix~\ref{apx:results}. 

\begin{table}[htb]
\centering
\begin{small}
\begin{tabularx}{\columnwidth}{lcc}
\toprule
\textbf{Evaluation Metric} & \textbf{Faith.} & \textbf{Comp.} \\ \midrule
SacreBLEU~\citep{Post2018ACF}    & 31.33  & 33.01 \\
ROUGE-L~\citep{Lin2003AutomaticEO}      & 27.69  & \underline{43.43} \\
METEOR~\citep{Banerjee2005METEORAA}       & 37.00  & 39.96 \\
BERTScore~\citep{Zhang2019BERTScoreET}    & 24.82  & 43.29 \\
A3CU~\citep{liu-etal-2023-towards-interpretable}        & 41.22 & 40.10 \\
TAPAS-Acc~\citep{Liu2022PLOGTP}    & -10.40 & 21.29 \\
G-Eval~\citep{Liu2023GEvalNE}       & \underline{47.82} & 26.35 \\
\rowcolor{gray!20}
\textbf{\eval} (ours) & \textbf{64.94} & \textbf{67.67} \\ \midrule
Inter-Human Correlation       & 84.81 & 75.70 \\
\bottomrule
\end{tabularx}
\vspace{-0.2cm}
\end{small}
\caption{Meta Evaluation results.}
\vspace{-0.2cm}
\label{tab:meta_main}
\end{table}

\definecolor{tab_gray}{HTML}{7d7d7d}

\begin{table*}[htb]
\setlength{\tabcolsep}{2pt}
\centering
\begin{small}
\renewcommand{\arraystretch}{1.0}

\begin{tabularx}{\textwidth}{
  l
  >{\centering\arraybackslash}X 
  >{\centering\arraybackslash}X 
  >{\centering\arraybackslash}X 
  >{\centering\arraybackslash}X 
  |>{\centering\arraybackslash}X 
  ||>{\centering\arraybackslash}X|  
  >{\centering\arraybackslash}X 
  >{\centering\arraybackslash}X 
  >{\centering\arraybackslash}X 
  >{\centering\arraybackslash}X 
  |>{\centering\arraybackslash}X 
  ||>{\centering\arraybackslash}X 
}

\toprule
\multirow{3}{*}{\textbf{Generator}} 
  & \multicolumn{6}{c}{\textbf{Faithfulness}} 
  & \multicolumn{6}{c}{\textbf{Completeness}} \\ 
\cmidrule(lr){2-7} 
\cmidrule(lr){8-13}
  & A\&S & C\&R & P\&O & T\&P & Avg.$\hat{T}_q$ &  Avg.$T_q$ 
  & A\&S & C\&R & P\&O & T\&P & Avg.$\hat{T}_q$ & Avg.$T_q$ \\ 
\midrule
 
\rowcolor{gray!20}
\multicolumn{13}{l}{\textit{Proprietary LLMs}} \\
\includegraphics[width=0.8em]{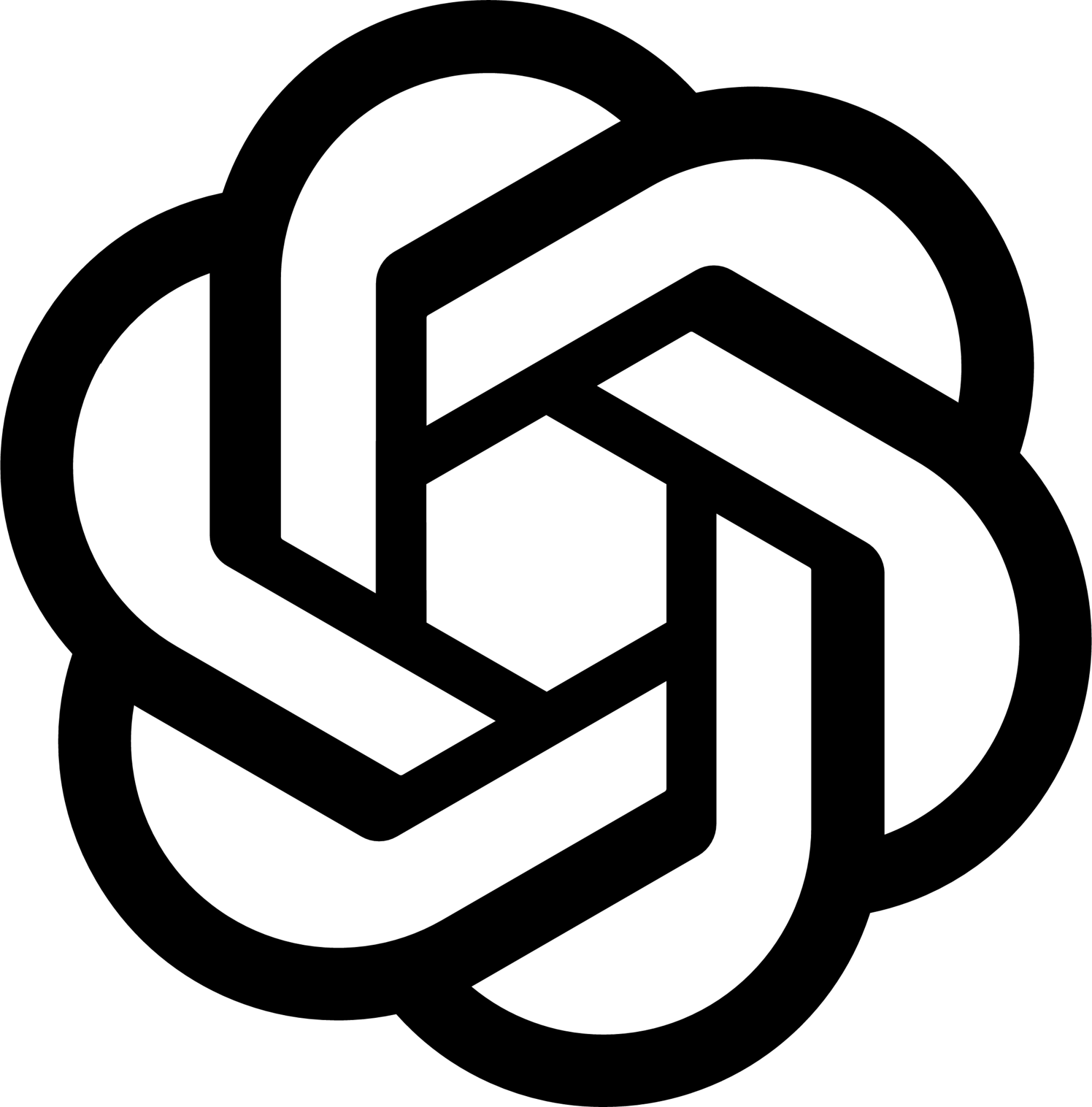} o3-mini 
  & \underline{38.62} & \underline{42.11} & \underline{35.12} & \underline{39.57} & \underline{38.85} & \textcolor{tab_gray}{42.57} 
  & 62.84 & \underline{58.78} & 59.77 & 59.42 & 60.20 & \textcolor{tab_gray}{60.81} \\
\includegraphics[width=0.8em]{sourceEMOJI/OpenAI.png} GPT-4o 
  & 36.60 & 39.79 & 33.96 & 37.57 & 36.98 & \textcolor{tab_gray}{41.46} 
  & \textbf{64.46} & \textbf{59.37} & \textbf{61.49} & \textbf{60.77} & \textbf{61.52} & \textcolor{tab_gray}{63.28} \\
\includegraphics[width=0.8em]{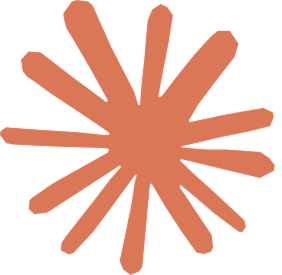} Claude 3.5 Sonnet 
  & \textbf{41.00} & \textbf{42.45} & \textbf{35.60} & \textbf{39.66} & \textbf{39.68} & \textcolor{tab_gray}{43.35} 
  & 61.67 & 56.06 & 58.94 & 57.86 & 58.63 & \textcolor{tab_gray}{59.70} \\
\midrule

\rowcolor{gray!20}
\multicolumn{13}{l}{\textit{Open-source LLMs}} \\
\includegraphics[width=0.8em]{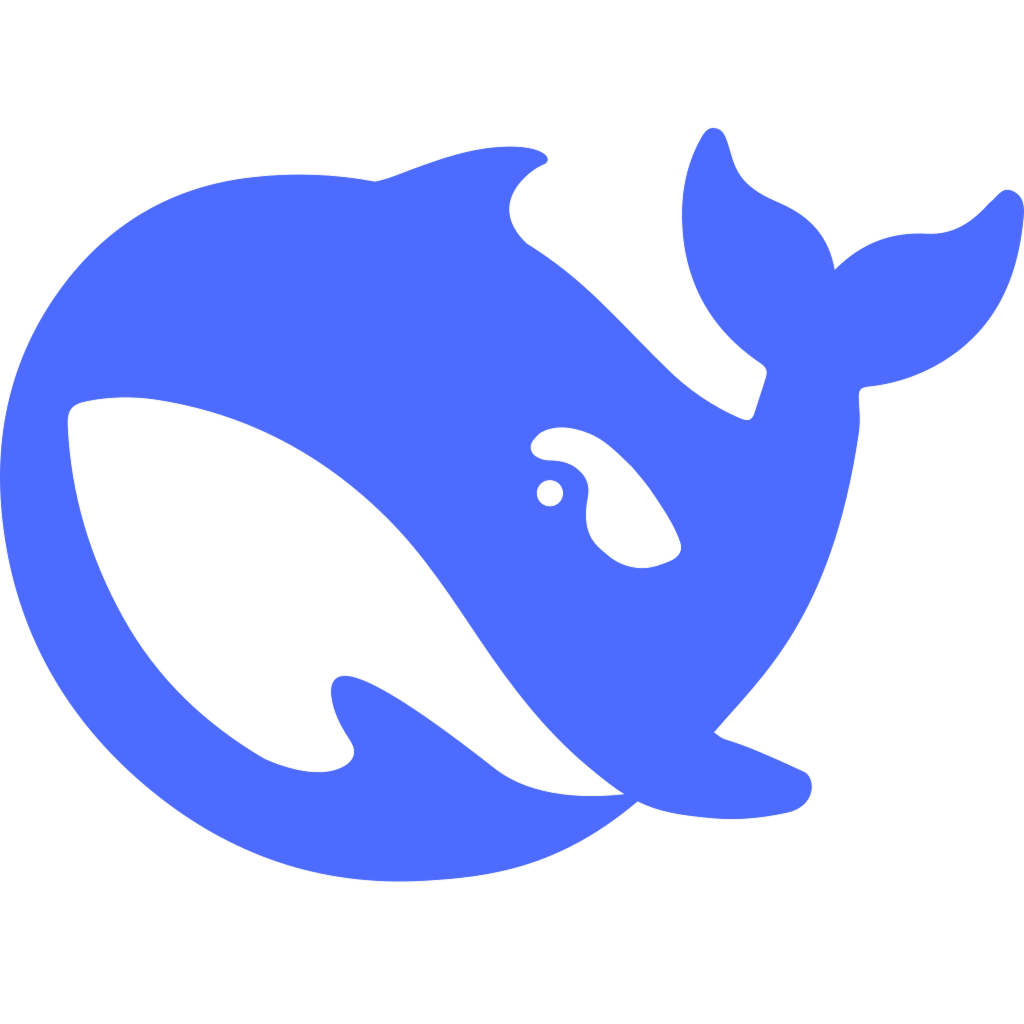} DeepSeek-R1-8B 
  & 37.05 & 36.57 & 34.09 & 34.51 & 35.55 & \textcolor{tab_gray}{40.12} 
  & \underline{64.23} & 58.72 & \underline{61.41} & \underline{59.48} & \underline{60.96} & \textcolor{tab_gray}{63.41} \\
\includegraphics[width=0.8em]{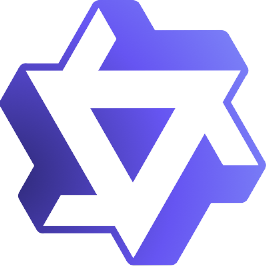} Qwen2-7B 
  & 33.79 & 36.97 & 29.85 & 33.01 & 33.40 & \textcolor{tab_gray}{40.59} 
  & 61.92 & 57.88 & 59.73 & 57.32 & 59.21 & \textcolor{tab_gray}{61.57} \\
\includegraphics[width=0.8em]{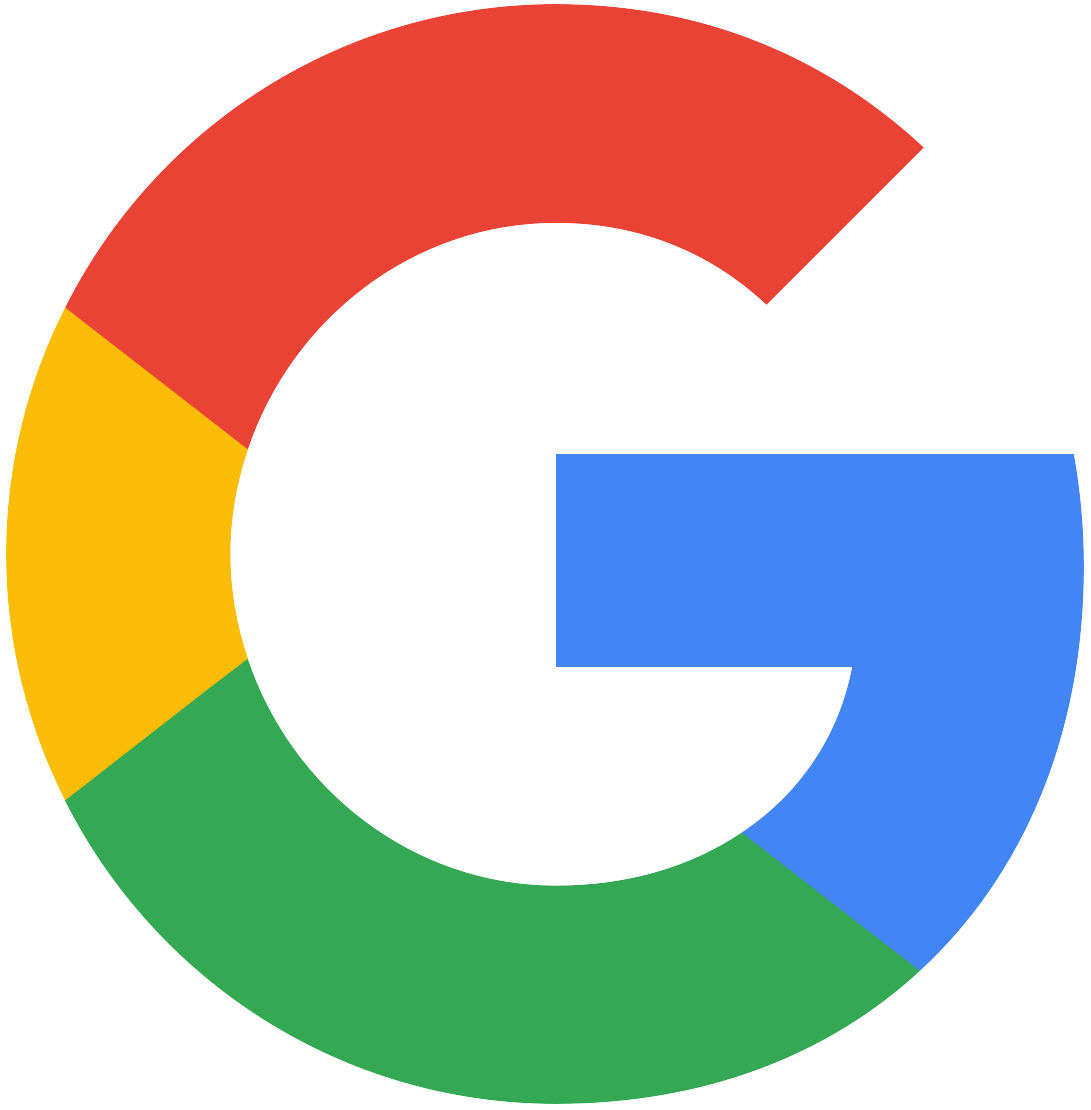} Gemma-7B 
  & 26.51 & 28.91 & 23.63 & 22.60 & 25.41 & \textcolor{tab_gray}{30.31} 
  & 62.72 & 57.73 & 60.52 & 59.68 & 60.16 & \textcolor{tab_gray}{61.64} \\
\includegraphics[width=0.8em]{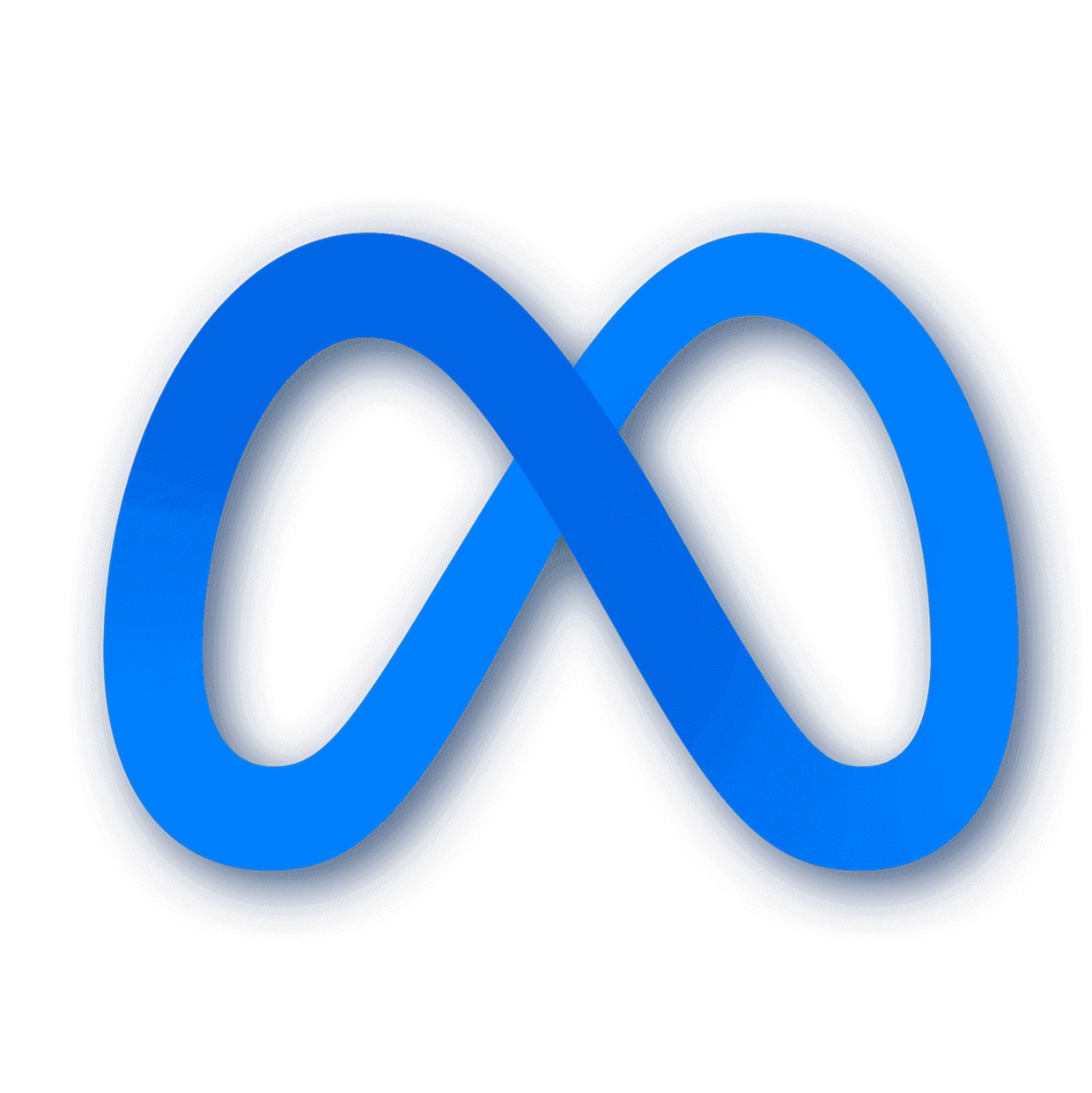} Llama 3.1-8B 
  & 32.30 & 33.71 & 29.73 & 30.63 & 31.59 & \textcolor{tab_gray}{37.62} 
  & 59.20 & 55.11 & 57.45 & 55.85 & 56.90 & \textcolor{tab_gray}{58.14} \\  
\includegraphics[width=0.8em]{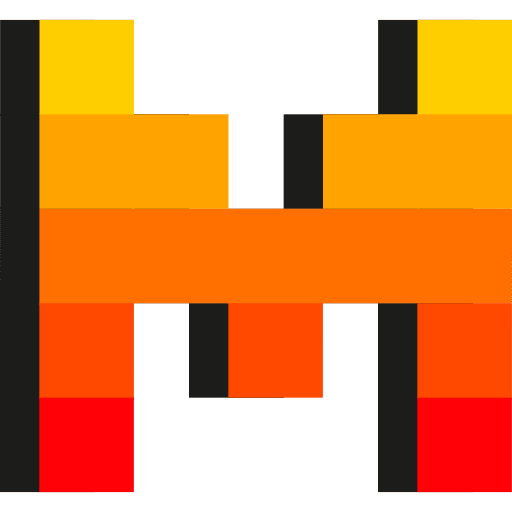} Mistral-7B 
  & 31.33 & 33.11 & 27.58 & 29.81 & 31.33 & \textcolor{tab_gray}{36.26} 
  & 62.96 & 58.00 & 59.66 & 58.29 & 59.73 & \textcolor{tab_gray}{62.77} \\
\midrule

\rowcolor{gray!20}
\multicolumn{13}{l}{\textit{SOTA TQA-methods}} \\
\includegraphics[width=0.8em]{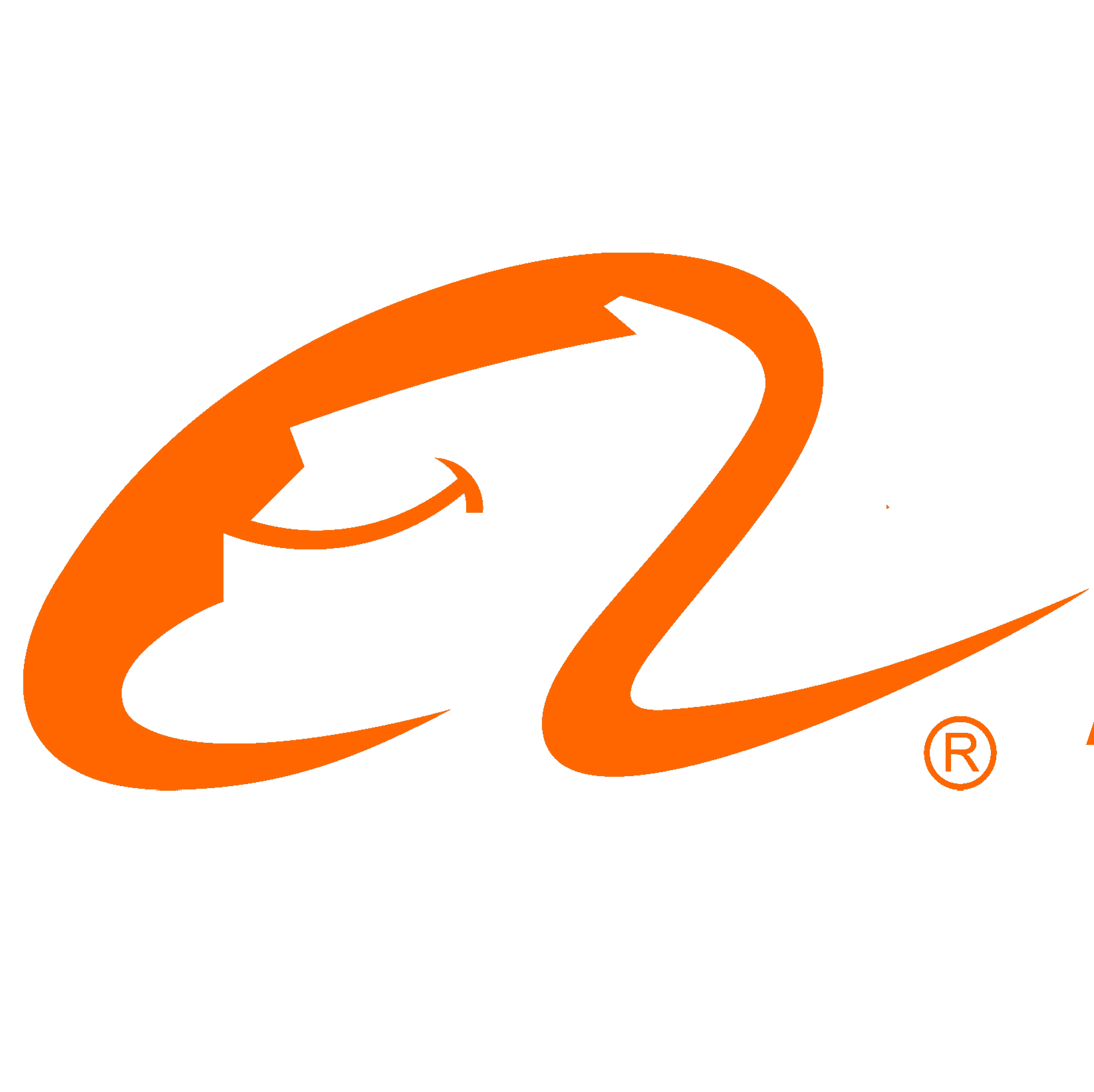} Dater 
  & 26.33 & 30.26 & 27.06 & 28.04 & 27.92 & \textcolor{tab_gray}{32.43} 
  & 59.70 & 57.26 & 60.35 & 59.16 & 59.12 & \textcolor{tab_gray}{62.32} \\
\includegraphics[width=0.8em]{sourceEMOJI/Google.png} Chain-of-Table 
  & 32.86 & 33.28 & 30.11 & 31.33 & 31.90 & \textcolor{tab_gray}{37.71} 
  & 59.29 & 55.49 & 58.58 & 57.17 & 57.63 & \textcolor{tab_gray}{62.15} \\
\includegraphics[width=0.8em]{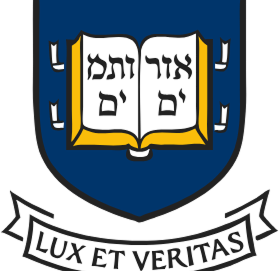} TaPERA 
  & 18.38 & 21.06 & 18.16 & 19.38 & 19.25 & \textcolor{tab_gray}{20.68} 
  & 59.23 & 53.56 & 58.05 & 55.53 & 56.59 & \textcolor{tab_gray}{55.28} \\
\bottomrule
\end{tabularx}
\end{small}

\caption{A\&S, C\&R, P\&O, and T\&P correspond to the types of question in Table~\ref{tab:statistics}. Avg.$\hat{T}_q$ denotes the average open-domain performance of different question types with retrieved tables, and Avg.$T_q$ indicates closed-domain performance using only the ground truth tables. Precision/Recall for Completeness are detailed in Appendix~\ref{apx:results}.}
\vspace{-0.2cm}
\label{tab:generator}
\vspace{-0.2cm}
\end{table*}

\begin{figure}[ht]
\centering
\includegraphics[width=\linewidth]{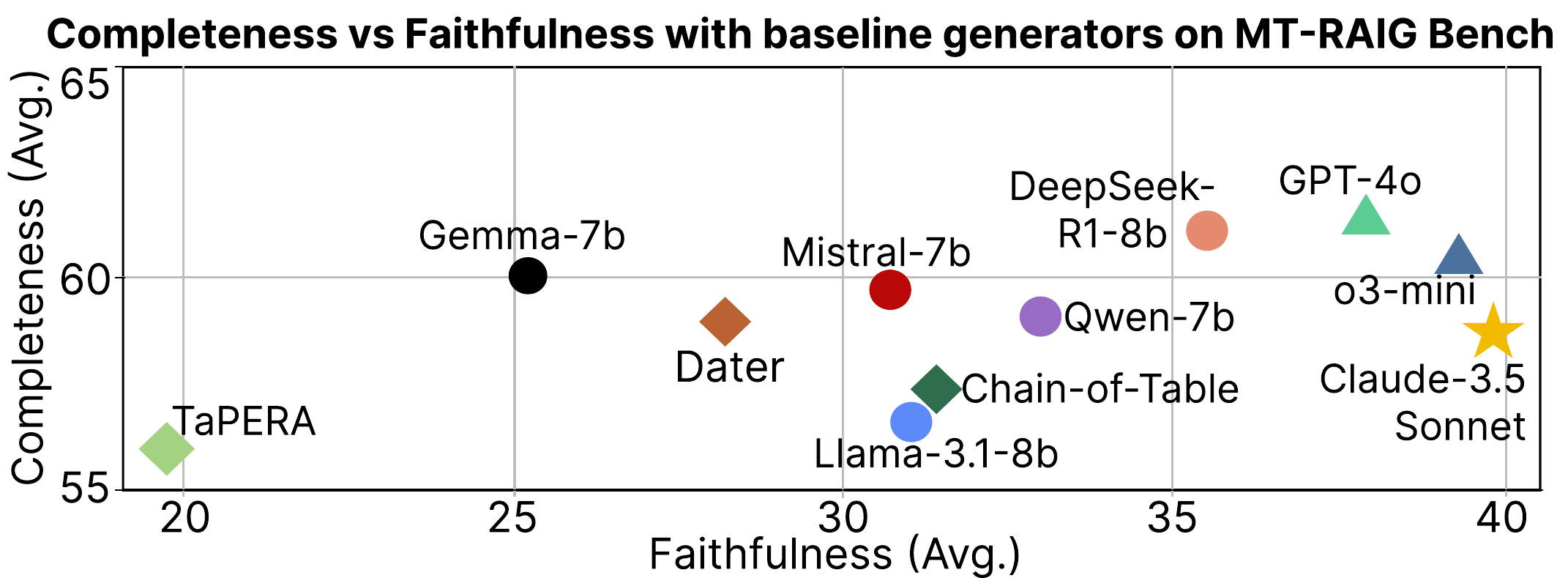}
\vspace{-0.4cm}
\caption{Comparsion of various baseline generators based on their average $\hat{T}_q$ completeness and faithfulness.}
\vspace{-0.5cm}
\label{fig:baselines_fig}
\end{figure}
\paragraph{Insight Generation Evaluation}
Leveraging the \eval, we evaluate the insight generation performance of various baselines on \bench. 
From the results in Table~\ref{tab:generator} and Figure~\ref{fig:baselines_fig}, we derive the following key conclusions:
\textbf{(1) \bench poses a significant challenge for insight-level table reasoning, even for frontier LLMs}. Both open-source and proprietary models struggle to generate insights for questions in \bench, achieving only around 40\% in faithfulness and 60\% in completeness even when provided with gold tables $T_q$ as input.
\textbf{(2) Deep thinking with scaling test-time compute also shows promise in insight-level table reasoning tasks.}
Although proprietary models generally outperform open-source counterparts, DeepSeek-R1-8b achieves performance on par with proprietary models and even surpasses Claude and o3-mini in completeness score, despite having a smaller number of parameters.
Given that a similar reasoning model, o3-mini, also demonstrates strong performance, these results supports  \citet{testtimecompute, DeepSeekAI2025DeepSeekR1IR}'s recent finding
that inference-time scaling substantially enhances models' capacity for complex reasoning, suggesting its applicability can also extend to insight-level table reasoning tasks. 
\noindent \textbf{(3) SOTA TQA-methods underperform compared to general-purpose LLMs.}
We attribute this to the focus of existing TQA methods on single-table factoid extraction and their
specialization in closed-domain scenarios, which limits their robustness to noise from irrelevant tables. 
\definecolor{fig_green}{HTML}{5dcc92}
\definecolor{fig_red}{HTML}{e58a6e}
\definecolor{fig_yellow}{HTML}{f1bc00}

\begin{figure}[ht]
\centering
\includegraphics[width=\linewidth]{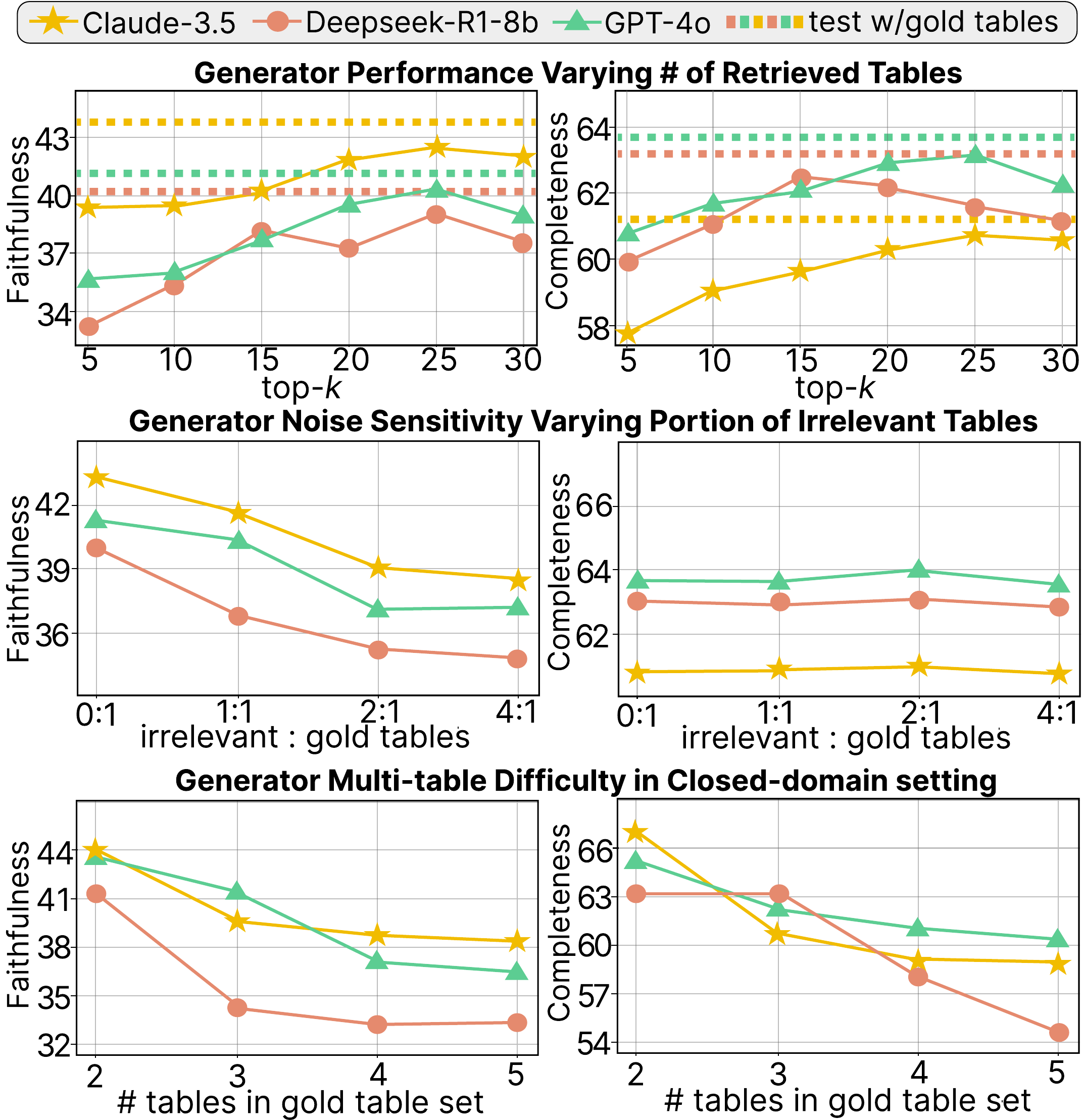}
\caption{\textbf{(Upper)} Generator performance varying the number of retrieved tables. Dash lines are closed-domain performance serving as the upper bound for each model. \textbf{(Middle)} Generator noise sensitivity varying portion of irrelevant tables against gold tables. \textbf{(Lower)} Generator multi-table difficulty comparison according to the number of tables in the gold table set.}
\label{fig:plot_exp}
\end{figure}

\subsection{Further Analysis}
\textbf{Trade-offs in retrieving more tables} \quad
We first analyze the relationship between retrieval and generation performance by varying the number of retrieved tables (\textit{k}).
While increasing the number of \textit{k} is an intuitive way to improve generation performance in RAG systems~\citep{kim2024is}, we investigate whether this also holds for the table domain, where retrieved information is structurally more complex than general text.
From the results in Figure~\ref{fig:plot_exp} (Upper), we first observe that the generation performance broadly aligns with conventional intuition, exhibiting an overall improvement across all models as \textit{k} increases.  
However, we also find that a continued increase in \textit{k} does not guarantee sustained performance gains. Beyond a certain threshold, performance plateaus or even declines, suggesting that retrieving an excessive number of tables introduces noise into the generation process.

\paragraph{Factuality of the generated insights are more sensitive to the noisy tables}
Guided by the observations in Figure~\ref{fig:plot_exp} (Upper), we further investigate how sensitively the generator responds to noisy table information. To isolate the impact of noise from retrieval errors, we analyze samples where the entire gold table set is successfully retrieved.
Within these samples, we analyze the performance by varying the ratio of irrelevant to gold tables while keeping the gold tables fixed as part of the model input.
The results in Figure~\ref{fig:plot_exp} (Middle) reveal that, as the proportion of irrelevant tables increases, faithfulness shows a marked decrease while completeness remains relatively stable. 
This suggests that although including the correct tables ensures the generator can somewhat address the high-level key steps (i.e., maintain completeness), the added noise from irrelevant tables hinders the extraction of precise factors for the final insights.

\paragraph{Even without the noisy tables, the model still struggles as the number of tables to reference increases}
While our findings indicate that generation performance degrades with noisy tables, one might wonder how challenging it becomes for the model as the number of source tables it should handle increases, even in noise-free conditions.
To explore this, we conduct a closed-domain experiment in which only the gold tables are provided as input, and investigate how generation performance varies depending on the number of tables in the gold table set that the model should reference.
From the results in Figure~\ref{fig:plot_exp} (Lower), we observe a significant drop in both faithfulness and completeness as the number of gold tables to reference grows, showing that current models struggle to conduct complex reasoning across multiple sources of tables.

\section{Related Work}
\label{sec:relatedWork}
\paragraph{Benchmarks for Table-based Tasks}
Benchmarks for table-based reasoning has been studied across several major tasks, including text-to-SQL~\citep{zhong2018seqsql}, table-to-text~\citep{Parikh2020ToTToAC}, and tableQA~\citep{pasupat-liang-2015-compositional}.
Early works~\citep{Yu2018SpiderAL,Nan2021FeTaQAFT} mainly focus on measuring the performance of neural models as executors for extracting factoid-level information, where models generate SQL queries or direct answers from tables.
More recently, some works~\citep{Zhao2023QTSummAN,Seo2024UnveilingIT} target insight-level table reasoning tasks, moving beyond extracting explicit information from tables.
Unlike prior works, our \bench significantly differs by exploring more realistic scenario where the system retrieves multiple evidence tables to generate insightful responses.
\paragraph{Automated Evaluation of Long-form Outputs in Table-based Tasks}
Existing methods have primarily compute the lexical overlap ~\citep{Post2018ACF}, or  semantic similarity with the reference~\citep{Zhang2019BERTScoreET}, while some~\citep{Liu2022PLOGTP} focus on factuality assessment using a trained verifier to check the grounding against the source table.
Some recent studies~\citep{zhao-etal-2024-tapera,wang-etal-2024-revisiting} find that LLM-based evaluation metrics (\textit{i.e.,} G-Eval), show stronger alignment with human evaluation.
Despite these advancements, these metrics still struggle to capture the fine-grained quality of long-form outputs. To address this gap, we propose a novel decomposition-based evaluation framework, \eval, which outperforms existing metrics in aligning with human judgments.

\section{Conclusion}
\label{sec:conclusion}
This work introduces \bench, the first large-scale benchmark for retrieval-augmented insight generation over multiple tables, alongside \eval, a novel automated evaluation framework designed to address the limitations of conventional metrics in assessing multi-table insights. Extensive experiments reveal that even frontier LLMs and SOTA TableQA systems struggle to meet the challenges posed by \bench.

\section*{Limitations}
\label{sec:limitation}
Although we believe that our \bench and \eval could serve as valuable resources for the research community, several limitations remain, suggesting areas for future improvement.
First, while the synthetic generation of \bench enables scalable and consistent dataset construction~\citep{tan2024largelanguagemodelsdata, hao2024syntheticdataaichallenges,han-etal-2024-rag} as we discussed in Section~\ref{sec:benchmark}, this approach carries inherent risks~\citep{10.1145/3442188.3445922}, such as reduced linguistic diversity in questions or potential overalignment with model generated responses that are unnatural or unfaithful description. 
Despite this downside, we minimize these risks by adopting a human-in-the-loop annotation process, where human experts iteratively refine machine-generated questions. Further, we conduct a machine-human dual quality control, in which automated filters first strictly discard low-quality instances, followed by human validation to ensure the quality of the final benchmark.

Second, although \bench covers both relational DB tables and Wikipedia tables that constitute the two main portion of table domain, it could be beneficial to add more data sources from diverse domains (e.g., financial, scientific, or medical tables).
Future works could examine how existing models perform in these specialized domains to assess their robustness to domain-specific terminology or extremely large table sets.

Lastly, a notable limitation lies in our \textsc{MT-RAIG Eval}’s reliance on LLM for assessment.
While recent works increasingly adopt LLM-based evaluators in diverse domains due to their scalability and flexibility~\citep{ye2024flask,kim2024stopplayingguessinggame, ru2024ragchecker, han-etal-2024-rag, wang-etal-2024-revisiting}, it is important to note that their judgments may still be affected by biases or inconsistencies inherent in the backbone model’s training data or architectural design.
However, we believe this issue has been largely minimized in our study. Through a meta-evaluation in Section~\ref{sec:evaluation}, we demonstrate our \textsc{MT-RAIG Eval}’s reliability in aligning with human judgments. Furthermore, additional experiments that replacing the \textsc{MT-RAIG Eval}'s LLM backbone with open-source models (Table~\ref{tab:apx_meta}) confirm the reproducibility of our evaluation framework.

\section*{Ethical Consideration}
\label{sec:ethical}
The output of text generation from LLMs may sometimes contain harmful, biased, or offensive content and contain the risk of potential hallucination~\citep{kim-etal-2024-verifiner}.
However, in our research, we assert that this risk is largely minimized.
The source tables and source natural language facts used in the construction of our \bench are derived from SPIDER~\citep{Yu2018SpiderAL} and Open-WikiTable~\citep{Kweon2023OpenWikiTableDF}, both of which are publicly available datasets licensed under the CC BY-SA 4.0, and these datasets have been annotated by human experts.
Additionally, we manually reviewed the generated texts and eliminate any toxic, offensive, or biased language to ensure the quality and fairness of the auto-generated contents. 
For human evaluation, two graduate students participated as annotators, receiving a pre-guide before taking part in evaluation and validation.
We advise the annotators to complete no more than 20 unit tasks per day, and the entire annotation process spanned approximately 30 days.

\section*{Acknowledgement}
\label{sec:ethical}
This work was supported by the IITP grants funded by the Korea government (MSIT) (No. RS-2020-II201361; RS-2024-00457882, AI Research Hub Project), and the NRF grant funded by the Korea government (MSIT) (No. RS-2025-00560295).

\bibliography{custom}

\newpage
\appendix

\section{Benchmark Construction Details}
\label{apx:benchmark}
\subsection{Data Annotation}
\label{apx:annotate_detail}
We show prompts used in data annotation process from Table \ref{pmt:step2} to \ref{pmt:step3_3}.
\paragraph{Multi-table Set Collection}
To extend the existing single table corpus into a collection of multi-table sets, we classify scenarios where multiple tables are combined into two categories: 
\begin{itemize}[leftmargin=*,topsep=4pt,itemsep=4pt,parsep=0pt]
\item\textit{Joinable tables}:
We first classify the SQL queries in the existing dataset SPIDER based on the join operator to identify tables that can be joined and then link the tables as the multi-table set. 

\item\textit{Topic-related tables}:
To cluster topically related tables, we semantically group the tables with the table meta data. Within each grouped table set, all tables are linked with related topics (at least two matching subtitles among page, section, or caption titles) and similar headers (differing by at most one column name). To consider the spatio-temporal relationships between tables, we use exact matching while excluding numeric values.
\end{itemize}

\paragraph{Question Types}
We provide the following definitions for each type of question to human reviewers who guide the agent with initial seed questions and validate the generated samples iteratively:
\begin{itemize}[leftmargin=*,topsep=4pt,itemsep=4pt,parsep=0pt]
    \item \textit{Analysis \& Summary}:
Synthesize multiple data sources into a coherent narrative, focusing on interpretation rather than raw figures. Identifies key metrics and contextualizes quantitative outcomes to uncover biases influencing decision-making.

    \item \textit{Performance \& Outcome}: 
Evaluate measurable achievements by linking performance metrics to qualitative success factors. Assesses growth and long-term accomplishments to provide a broader context for understanding progress.
    
    \item \textit{Comparison \& Relationship}:
Analyze relationships between data points, explaining connections between attributes. Examines structural components and grouping logic to reveal organizational hierarchies and patterns, enhancing understanding of data organization.
    \item \textit{Trend \& Pattern}:
Examine temporal or categorical changes to identify recurring behaviors and systemic shifts. Goes beyond documenting events to interpret transformations, providing insights into structural changes or cyclical phenomena.
\end{itemize}

\subsection{Quality Control}
\label{apx:verify_detail}
\paragraph{Agent Self-Verification}
We first utilize an agent annotator as a self-verifier, ensuring that each multi-table set, question, and insight triple satisfies strict criteria:

\begin{itemize}[leftmargin=*,topsep=4pt,itemsep=4pt,parsep=0pt]
    \item \textbf{\textit{Relevance}}: Does the question appropriately capture the relationships between the tables in the multi-table set, ensuring that it pertains to all tables and can be answered solely using the provided information?
    
    \item \textbf{\textit{Faithfulness}}: Does the insight accurately reflect the information within the multi-table set, ensuring it remains grounded in the given data while providing a clear and unambiguous response?
    
    \item \textbf{\textit{Completeness}}: Does the insight fully and logically address the question, covering all necessary aspects while maintaining clarity and coherence?
\end{itemize}

Any triple (multi table set, question, and insight) that fails at least one of these standards is discarded.
Representative examples of discarded data are in Table \ref{case:filter_i} and \ref{case:filter_ii}.
We report the statistics of self-verification process in Table \ref{tab:apx_ratio}.
Prompts used in agent verification process for three criteria are shown in Table \ref{pmt:step4_1}.
\begin{table}[htb]
\centering
\begin{small}
\renewcommand{\arraystretch}{1.0}

\begin{tabularx}{\linewidth}{%
>{\centering\arraybackslash}p{0.16\linewidth}
>{\centering\arraybackslash}X
>{\centering\arraybackslash}X
>{\centering\arraybackslash}X}  
\toprule
\textbf{Ratio} & \textbf{Relevance} & \textbf{Faithfulness} & \textbf{Completeness} \\
\midrule

\% & 64.73 & 19.38 &  7.47 \\

\bottomrule
\end{tabularx}
\end{small}
\caption{Discarded data ratio for each criterion in self-verification process. Only triples that pass all these standards are used as \bench.}
\label{tab:apx_ratio}
\end{table}

\begin{figure*}[ht]
\centering
\includegraphics[width=\textwidth]{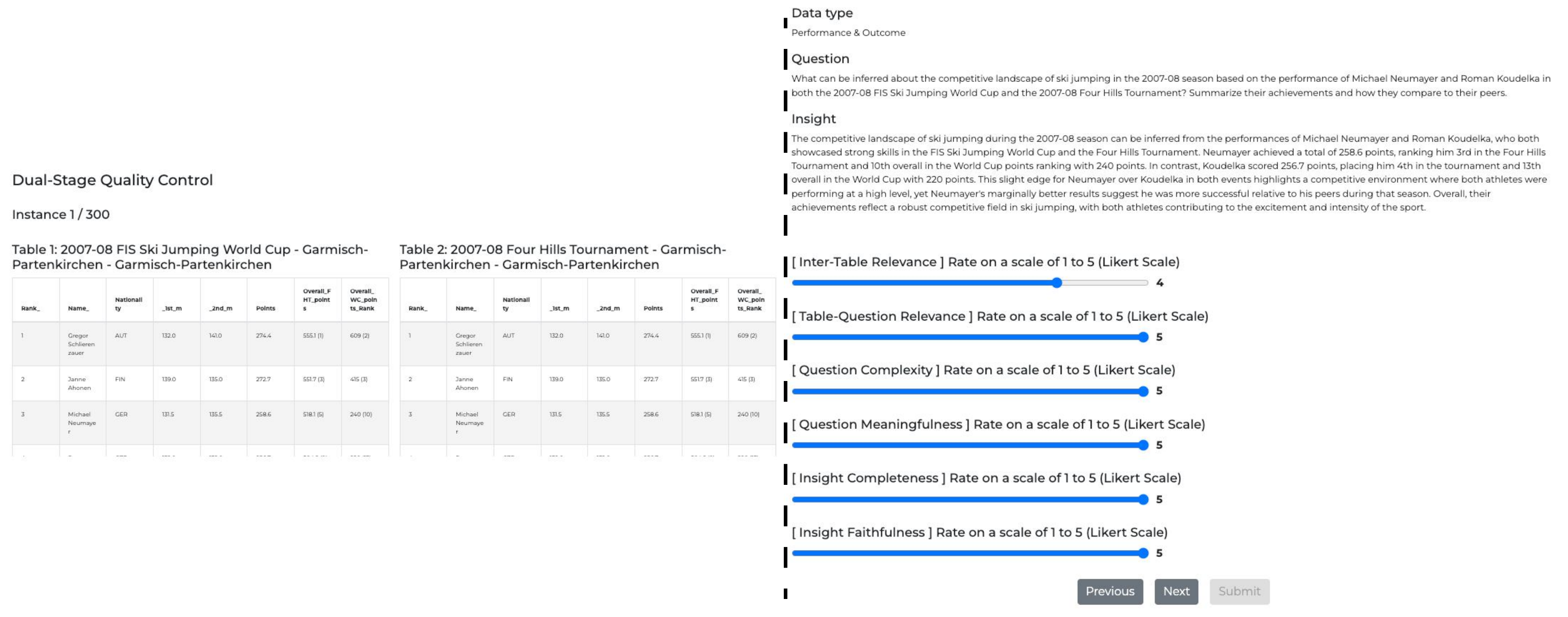}
\caption{Human validation interface.}
\label{fig:step4_2}
\end{figure*}

\paragraph{Human Validation}
Alongside the agent verification, we incorporate human validation as critical review points for the annotated data, balancing the efficiency of automated annotation with the reliability of human annotation to ensure the quality of MT-RAIG BENCH.
We provide the human validation interface in Figure \ref{fig:step4_2}.
Specifically, we adopt the following criteria to check the quality of table,question,and insight triples:
\begin{itemize}[leftmargin=*,topsep=4pt,itemsep=4pt,parsep=0pt]
\item \textbf{\textit{Inter-Table Relevance}}: Measures how effectively data from different tables can be connected or combined to provide comprehensive insights.
    
 \item \textbf{\textit{Table-Question Relevance}}:Assesses the extent to which the contents of a table directly support and address the given question.
  \item \textbf{\textit{Quesiton Complexity }}:Evaluates the level of difficulty and the number of factors or layers involved in understanding or answering the question.
   \item \textbf{\textit{Question Meaningfulness}}:Determines whether the question is clearly defined, significant, and natural.
\item \textbf{\textit{Question-Insight Completeness}}: Checks if the insight provides all the necessary key information to address the question.
     \item \textbf{\textit{Table-Insight Faithfulness}}:Ensures that the insights drawn accurately and reliably reflect the source tables.
\end{itemize}

\section{Experimental Details}
\label{apx:experiments}
\subsection{Meta Evaluation}
\label{apx:setup}
\begin{figure*}[ht]
\centering
\includegraphics[width=\textwidth]{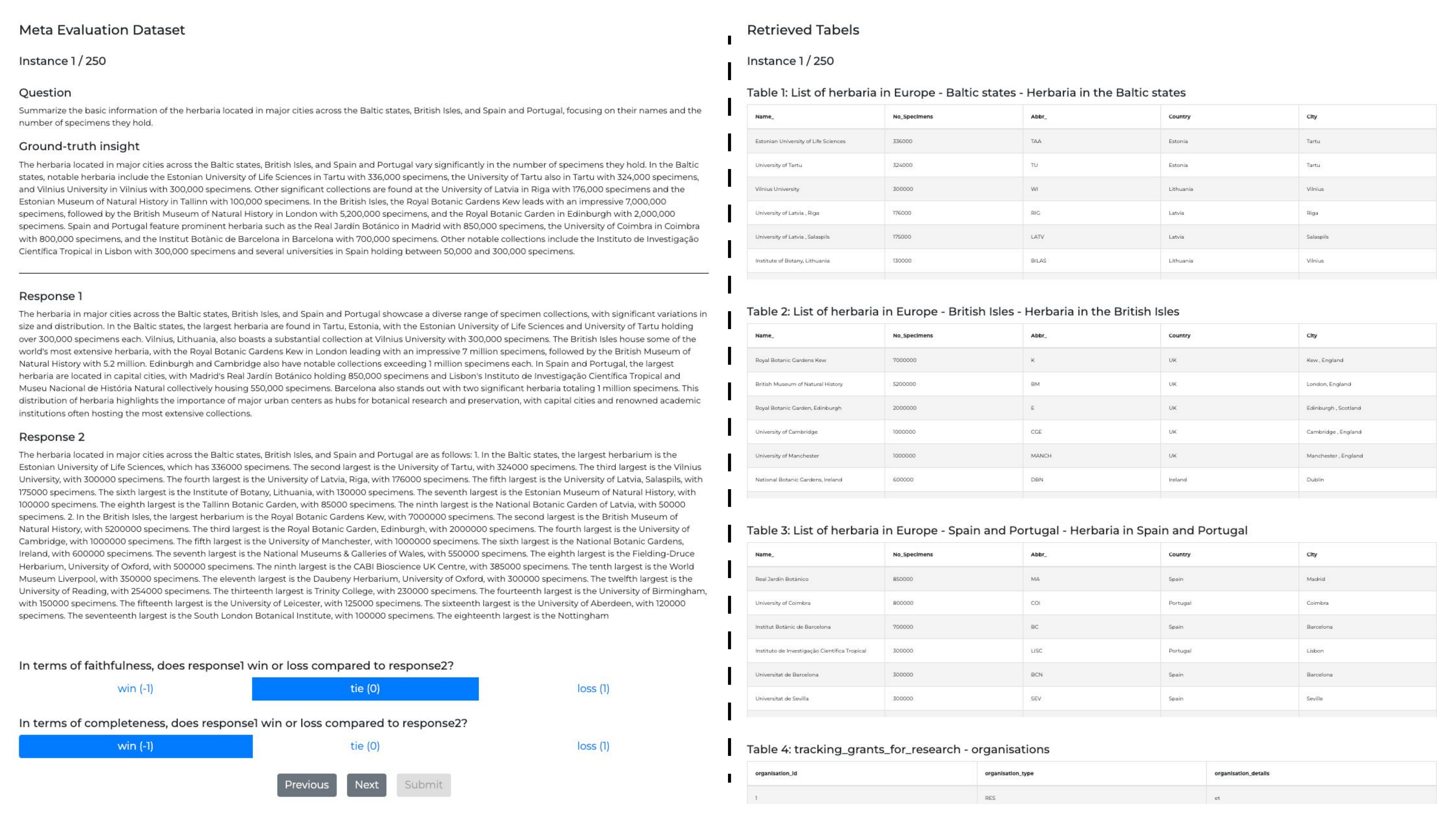}
\caption{Human annotation interface of the meta evaluation dataset.}
\label{fig:meta_ui}
\end{figure*}

Considering that all automatic metrics are designed with distinct objectives and functionalities, direct numerical comparisons between their scores are inherently limited. 
Instead, it is intuitive that a metric’s reliability should be judged by its capacity to align with relative human preferences. 
To operationalize this principle, we adopt the meta-evaluation protocol proposed by \citet{ru2024ragchecker}, constructing a meta evaluation dataset of 250 response pairs sampled from 11 different baseline generators.
Human evaluators are then tasked with annotating each pair across two dimensions—faithfulness and completeness—by selecting one of three options: win, tie, or loss.
The annotation interface is illustrated in Figure \ref{fig:meta_ui}.
We implement \eval on the top of \texttt{gpt-4o-mini-2024-07-18}. 
To check the reproducibility of \eval, we additionally conduct an ablation study in Table~\ref{tab:apx_autoeval}, examining the impact of different backbone LLMs on \eval.
 We consider the following automatic metrics as baselines:

\begin{itemize}[leftmargin=*,topsep=4pt,itemsep=4pt,parsep=0pt]

    \item\textbf{SacreBLEU}~\citep{Post2018ACF} standardizes BLEU ~\cite{Papineni2002BleuAM} score calculations by ensuring consistent and reproducible results.
    It measures the geometric mean of n-gram precision over the output text.
    
    \item\textbf{ROUGE-L}~\citep{Lin2003AutomaticEO} evaluates text similarity based on the longest common subsequence.
    Specifically, we reported F1 score.
    
    \item\textbf{METEOR}~\citep{Banerjee2005METEORAA} evaluates text similarity by using unigram matching between machine-generated outputs and human reference texts.
    
    \item\textbf{BERTScore}~\citep{Zhang2019BERTScoreET} computes the similarity between generated and reference texts using contextual word embeddings.
    
    \item\textbf{A3CU}~\citep{liu-etal-2023-towards-interpretable} evaluates summarization quality by directly comparing texts without extracting atomic content units, providing a human-aligned assessment of content similarity.
    
    \item\textbf{TAPAS-Acc}~\citep{Liu2022PLOGTP} is a reference-free metric that uses a TAPAS~\citep{Herzig2020TaPasWS} model fine-tuned on the TabFact~\cite{Chen2019TabFactAL} dataset to assess the faithfulness of generated text by verifying factual consistency.
    
    \item\textbf{G-Eval}~\citep{Liu2023GEvalNE} assesses the quality of generated text based on specific evaluation criteria using LLMs.
    We adopt G-Eval to separately evaluate \textit{faithfulness} and \textit{completeness} on a 5-point Likert scale.
    The evaluation prompts are provided in Table \ref{pmt:apx_geval}.

\end{itemize}


\subsection{Retrieval Baselines}
\label{apx:rtr_base}
To assess the effectiveness of multi-table retrieval, we evaluate both general-purpose retrievers and table-specific retrievers.
General-purpose retrievers serve as strong baselines, given their widespread use in plain retrieval tasks.
Table-specific retrievers are designed explicitly for tables, leveraging table-specific representations for improved retrieval performance.
Specifically, we consider the following retrievers as the baselines:
\paragraph{General-purpose retrievers}
\begin{itemize}[leftmargin=*,topsep=4pt,itemsep=4pt,parsep=0pt]

    \item\textbf{BM25}~\citep{Robertson2009ThePR} is a sparse retriever that relies on a traditional bag-of-words representation to score relevant documents.
    In this study, we rank linearized tables as documents based on term frequency and inverse document frequency, leveraging lexical overlaps between questions and tables.

    \item\textbf{DPR}~\citep{Karpukhin2020DensePR} employs BERT-based encoders to independently map questions and documents into a shared embedding space.
    It learns dense vector representations that enable semantic similarity matching.
    In this study, we implement DPR utilizing \texttt{facebook/dpr-} \texttt{encoder-multiset-base}.

    \item\textbf{Contriever}~\citep{Izacard2021UnsupervisedDI} encodes questions and documents into a shared embedding space, optimizing for relevance through contrastive learning.
    In this study, we implement Contriever utilizing \texttt{facebook/} \texttt{contriever-msmacro}.

\end{itemize}

\paragraph{Table-specific retrievers}
\begin{itemize}[leftmargin=*,topsep=4pt,itemsep=4pt,parsep=0pt]

    \item\textbf{DTR}~\citep{Herzig2021OpenDQ} is a table-specific dense retriever built on a TAPAS~\citep{Herzig2020TaPasWS} backbone, designed to effectively encode tabular structures and relationships.
    In this study, we fine-tune \texttt{google/tapas-base}.

    \item\textbf{TableLlama}~\citep{Zhang2023TableLlamaTO} is Llama 2-7B~\citep{Touvron2023Llama2O} based open-source LLM-based generalist model that is designed for a variety of table-based tasks.
    We utilize TableLlama to generate question and table embeddings for table retrieval task.
    In this study, we fine-tune \texttt{osunlp/TableLlama}.

\end{itemize}

\subsection{Generator Baselines}
\label{apx:gen_base}
To comprehensively compare the performance of insight generation across multiple tables, we consider three types of baselines: proprietary LLMs, open-source LLMs, and SOTA table question-answering methods. 
Proprietary LLMs encompass commercial models renowned for their advanced reasoning and high-quality text generation capabilities. Open-source LLMs, serving as transparent and adaptable alternatives, leverage large-scale training data and advanced architectures to deliver performance that is competitive with proprietary counterparts. SOTA TQA methods are specialized approaches explicitly engineered for complex, table-aware reasoning tasks, prioritizing accuracy in tabular question-answering. 
We consider the following models as the baseline generators: 

\paragraph{Proprietary LLMs}
\begin{itemize}[leftmargin=*,topsep=4pt,itemsep=4pt,parsep=0pt]

    \item\textbf{o3-mini}~\citep{O3mini} is a smaller yet advanced LLM developed by OpenAI, which is designed to efficiently solve complex problems by breaking them into constituent parts.
    In this study, we leverage \texttt{o3-mini-2025-01-31} with reasoning effort parameter as medium.

    \item\textbf{GPT-4o}~\citep{GPT4o} is an advanced OpenAI's proprietary LLM known for its enhanced reasoning capabilities and performance across various disciplines.
    In this study, we leverage \texttt{gpt-4o-2024-08-06} checkpoint.

    \item\textbf{Claude 3.5 Sonnet}~\citep{Claude3S} is developed by Anthropic.
    It features improvements in coding proficiency and multimodal capabilities.
    In this study, we leverage \texttt{claude-3-5-sonnet} \texttt{-20241022} checkpoint.

\end{itemize}

\paragraph{Open-source LLMs}
\begin{itemize}[leftmargin=*,topsep=4pt,itemsep=4pt,parsep=0pt]

    \item\textbf{DeepSeek-R1-8B}~\citep{DeepSeekAI2025DeepSeekR1IR} is distilled version of DeepSeek-R1, which is a open-source LLM released by DeepSeek AI.
    In this study, we utilize \texttt{unsloth/DeepSeek-R1-Distill-Llama} \texttt{-8B}, which is based on Llama-3.1-8B.

    \item\textbf{Qwen2-7B}~\citep{Yang2024Qwen2TR} is an open-source LLM developed by Alibaba Cloud, with the largest model containing 72 billion parameters.
    In this study, we utilize \texttt{Qwen/Qwen2-7B-} \texttt{Instruct}.

    \item\textbf{Gemma-7B}~\citep{Mesnard2024GemmaOM} is an open-source LLM developed by Google DeepMind, which is known for its multilingual capabilities and creative outputs.
    In this study, we utilize \texttt{google/gemma-7b-it}.

    \item\textbf{Llama 3.1-8B}~\citep{llama-3} is an open-source LLM released by Meta AI, which is available in multiple sizes up to 405 billion parameters.
    In this study, we utilize \texttt{meta-llama/Llama-} \texttt{3.1-8B-Instruct}.

    \item\textbf{Mistral-7B}~\citep{Jiang2023Mistral7} is an open-source LLM developed by Mistral AI.
    In this study, we utilize \texttt{mistralai/Mistral-7B-} \texttt{Instruct-v0.3}.

\end{itemize}

\paragraph{SOTA TQA-methods}
\begin{itemize}[leftmargin=*,topsep=4pt,itemsep=4pt,parsep=0pt]

    \item\textbf{Chain-of-Table}~\citep{Wang2024ChainofTableET} is LLM-based method designed to enhance table-based reasoning.
    It employs iterative reconstruction of input table through dynamic tabular operations.

    \item\textbf{TaPERA}~\citep{zhao-etal-2024-tapera} a modular framework designed to enhance faithfulness and interpretability in long-form table question answering by combining a QA-based content planner and execution-based reasoning..

    \item\textbf{Dater}~\citep{Ye2023LargeLM} focuses on selecting relevant information from input tables and providing contextual information to support the statement verification process.

\end{itemize}

\subsection{Implementation Details}

\paragraph{Table Input Serialization}
Following recent studies that utilize language models for table-related tasks~\citep{chen2023largelanguagemodelsfew1shot,Seo2024UnveilingIT}, we serialize the table input into a flattened sequence to effectively represent table data for language model processing.
The table title is enclosed within \texttt{[TITLE]} tags, followed by the table headers marked with \texttt{[HEADER]}, where individual column names are separated by a vertical bar (\texttt{|}).
Each row is prefixed with a \texttt{[ROW]} tag and an index, while cell values are separated by a vertical bar.
This approach ensures that the table format is preserved while making the input compatible with language models.
For example, the input table is formatted as follows:

\begingroup
\spaceskip=3pt
\texttt{[TITLE] title [HEADER] col 1 | col2 | ... [ROW 1] cell 1,1 | cell 1,2 | ... [ROW 2]  cell 2,1 | ...}
\endgroup

\paragraph{Training and Inference}
For the table-specific retriever, we fine-tune each model on a single-table QA dataset using the AdamW optimizer with a learning rate of for 5 epochs. To efficiently finetune the model, we adopt LoRA and set the parameters as $r$ = 8, $\alpha$ = 32.  We use a constant learning rate schedule set at 2e-5, and train with the batch size of 1 on 4 NVIDIA A100 GPU. The inference of open-source LLMs are conducted using vLLM framework~\citep{kwon2023efficientmemorymanagementlarge}.  We set 
temperature to 0.0 for efficient and robust output generation.

\section{Ablation Study}
\label{apx:ablation}
\subsection{Reproducibility of \eval Backbone}
To assess whether \eval produces consistent results across different backbone LLMs, we analyze two open-source models (DeepSeek-R1-8B and Llama-3.1-8B) against the GPT-4o-mini, which is the original backbone of \eval. Specifically, we measure how closely the evaluation scores from each open-source model align with those from GPT-4o-mini by calculating pairwise correlation.
From the results in Table~\ref{tab:apx_evalbaseline}, we can observe that even when replacing the backbone with open-source models, the high correlation persists, demonstrating that our evaluation method is both reproducible and robust.

\begin{table}[htb]
\centering
\begin{small}
\renewcommand{\arraystretch}{1.0}

\begin{tabularx}{\linewidth}{%
l
>{\centering\arraybackslash}X
>{\centering\arraybackslash}X}  

\toprule
\textbf{Backbone} & \textbf{Faithfulness}& \textbf{Completeness} \\ \midrule

DeepSeek R1-8B     & 74.12 & 77.85 \\
Llama-3.1-8B       & 80.78 & 72.82 \\

\bottomrule
\end{tabularx}
\end{small}
\caption{Pearson correlation scores varying \eval backbone compared to GPT-4o mini.}
\label{tab:apx_evalbaseline}
\end{table}

\subsection{Effect of Parameter Scaling on \bench Performance}
From the results in Table~\ref{tab:generator}, we observe that proprietary models with larger model sizes generally show higher performance compared to open-source models that have relatively fewer parameters. 
To further investigate this, we conduct additional experiments by differ the parameter sizes of open-source LLM to understand the effect of model size in the table reasoning performance.
Specifically, we evaluate DeepSeek-R1, which is most powerful open-source LLM among our baselines ranging from 8B to 70B parameters.
From the results in Table~\ref{tab:apx_deepseek}, we confirm a general trend where increasing model size correlates with improved performance, with a particularly notable improvement in the faithfulness score.
This discrepancy suggests that the model’s capacity to accurately interpret and reason about table structures (which is a prerequisite for faithfulness) is more directly enhanced by parameter scaling, compared to the broader task coverage implied by completeness.
\definecolor{tab_gray}{HTML}{7d7d7d}
\newcommand{\downlogo}[2][{-1em}]{%
  \raisebox{#1}{\includegraphics[width=0.8em,keepaspectratio]{#2}}%
}

\begin{table}[htb]
\centering
\begin{small}
\renewcommand{\arraystretch}{1.0}

\begin{tabularx}{\linewidth}{%
l
>{\centering\arraybackslash}X
>{\centering\arraybackslash}X
>{\centering\arraybackslash}X
>{\centering\arraybackslash}X}  

\toprule
\multirow{2.5}{*}{\textbf{DeepSeek-R1}} 
& \multicolumn{2}{c}{\textbf{Faithfulness}}
& \multicolumn{2}{c}{\textbf{Completeness}} \\
\cmidrule(lr){2-3} \cmidrule(lr){4-5}
& \textbf{Score} & \textbf{$\Delta$\%} & \textbf{Score} & \textbf{$\Delta$\%} \\
\midrule

Distill-Llama-8B  & 35.55 & \textcolor{tab_gray}{-} & 60.96 & \textcolor{tab_gray}{-} \\ \midrule
Distill-Qwen-14B  & 37.26 & \textcolor{tab_gray}{+ 4.81} & 61.07 & \textcolor{tab_gray}{+ 0.18} \\
Distill-Qwen-32B  & \textbf{39.02} & \textcolor{tab_gray}{+ 9.76} & 60.62 & \textcolor{tab_gray}{\textminus\;0.56} \\
Distill-Llama-70B & 38.11 & \textcolor{tab_gray}{+ 7.20} & \textbf{62.13} & \textcolor{tab_gray}{+ 1.92} \\

\bottomrule
\end{tabularx}
\end{small}
\caption{\eval results varying parameter size of DeepSeek-R1. $\Delta$\% denotes the relative improvement in performance compared to the Distill-Llama-8B.}
\label{tab:apx_deepseek}
\end{table}

\section{Detailed Experimental Results}
\label{apx:results}
To supplement Table \ref{tab:meta_main},  we present the Spearman correlation results alongside the Pearson correlation, in terms of faithfulness and completeness, as shown in Table \ref{tab:apx_meta}.
We report the full multi-table retrieval results in Table \ref{tab:apx_rtr}, \eval Completeness score results of diverse baselines in Table \ref{tab:apx_comp}, with respect to Precision, Recall, and F1 score across different top-\textit{k} values.
Additionally, we present insight generation results along with existing automatic metrics in Table \ref{tab:apx_autoeval}.

\section{Case Study}
\label{apx:case_study}
We select representative examples of (1) \bench in Table \ref{case:bench_i} and \ref{case:bench_ii}, (2) discarded data in Table \ref{case:filter_i} and \ref{case:filter_ii}, (3) \eval Faithfulness and (4) Completeness evaluation in Table \ref{case:eval_faith} and \ref{case:eval_comp}, respectively.
Moreover, we present (5) comparison with \eval results for each baseline in Table \ref{case:comparison_i} and \ref{case:comparison_ii}.
Ensure that we display only the first three rows of each table to improve readability.

\begin{table*}[htbp]
\centering
\begin{tabularx}{\textwidth}{X}
\toprule
\textbf{Question Annotation} \\ \midrule
\textcolor{blue}{\textbf{[Instruction]}} \\
You are a helpful assistant that generate questions based on keywords derived from the table titles and overlapping values that contain information about relationships between tables. Please refer to the given examples to capture the insight-level pattern, and generate a total of 10 questions. \\\\
\textcolor{blue}{\textbf{[Inputs]}} \\
\textbf{Examples}: \{seed\_questions\} \\
\textbf{Keywords}: \{key\_words\_derived\_from\_table\_titles\} \\
\textbf{Overlapping values}: \{overlapping\_values\} \\\\
\textcolor{blue}{\textbf{[Output]}} \\
\textbf{Questions}: \\ \bottomrule
\end{tabularx}
\caption{Prompt used for question annotation process.}
\label{pmt:step2}
\end{table*}
\begin{table*}[htbp]
\centering
\begin{tabularx}{\textwidth}{X}
\toprule
\textbf{Multi-Table Facts Expansion} \\ \midrule
\textcolor{blue}{\textbf{[Instruction]}} \\
You are a helpful assistant that write a Python code to extract enriched facts from the given natural language facts based on table schemas. Please write a Python code in the following format. \\

\begin{lstlisting}[label={lst:code}]
```python
import pandas as pd # can import additional python libraries if necessary

def extract_patterns(dataframe, source_facts):
    ###CODE###
    return patterns

def expand_facts(dataframe, patterns):
    """Expand facts

    [Param]
    dataframe : pd.DataFrame
    patterns  : List[str]

    [Return]
    expanded_facts_list : List[str]
    """
    ###CODE_WITH_ITERATION###

    return expanded_facts_list
```
\end{lstlisting}

\textcolor{blue}{\textbf{[Inputs]}} \\
\textbf{Table schemas}: \{table\_schemas\} \\
\textbf{NL facts}: \{natural\_language\_facts\} \\\\
\textcolor{blue}{\textbf{[Output]}} \\
\textbf{Python code}: \\ \bottomrule
\end{tabularx}
\caption{Prompt used for multi-table facts expansion process.}
\label{pmt:step3_1}
\end{table*}
\begin{table*}[htbp]
\centering
\begin{tabularx}{\textwidth}{X}

\toprule
\textbf{Question-Relevant Knowledge Extraction} \\ \midrule
\textcolor{blue}{\textbf{[Instruction]}} \\
You are a helpful assistant that extract knowledge from natural language context that are relevant to the given question. Please filter out irrelevant content and extract question-relevant knowledge. \\\\
\textcolor{blue}{\textbf{[Inputs]}} \\
\textbf{Question}: \{question\} \\
\textbf{NL context}: \{expanded\_natural\_language\_facts\} \\\\
\textcolor{blue}{\textbf{[Output]}} \\
\textbf{Question-relevant knowledge}: \\ \midrule
\end{tabularx}
\caption{Prompt used for question-relevant knowledge extraction process.}
\label{pmt:step3_2}
\end{table*}
\begin{table*}[htbp]
\centering
\begin{tabularx}{\textwidth}{X}
\toprule
\textbf{Insight Annotation} \\ \midrule
\textcolor{blue}{\textbf{[Instruction]}} \\
You are a helpful assistant that generate an insight to answer the given question using knowledge. Please generate an insight. \\\\
\textcolor{blue}{\textbf{[Inputs]}} \\
\textbf{Question}: \{question\} \\
\textbf{Knowledge}: \{question\_relevant\_knowledge\} \\\\
\textcolor{blue}{\textbf{[Output]}} \\
\textbf{Insight}: \\ \bottomrule
\end{tabularx}
\caption{Prompt used for insight annotation process.}
\label{pmt:step3_3}
\end{table*}
\begin{table*}[htbp]
\centering
\begin{tabularx}{\textwidth}{X}
\toprule

\textbf{Self-Verification : Relevance} \\ \midrule
\textcolor{blue}{\textbf{[Instruction]}} \\
You are a strict verifier for evaluate relevance. Please verify whether the given multi-table set and question pair is relevant based on the given criteria. \\\\
\textbf{Criterion 1}: Does the question focus on the relationships between the tables in the multi-table set? \\
\textbf{Criterion 2}: Does the question relate to all the tables in the multi-table set? \\
\textbf{Criterion 3}: Does the question rely solely on the information in the multi-table set for an answer? \\
\textbf{Criterion 4}: Does the question avoid requiring information that is not present in the multi-table set? \\\\
\textcolor{blue}{\textbf{[Inputs]}} \\
\textbf{Multi-table set}: \{serialized\_tables\} \\
\textbf{Question}: \{question\} \\\\
\textcolor{blue}{\textbf{[Output]}} \\
\textbf{Verification}: \\

\midrule

\textbf{Self-Verification : Faithfulness} \\ \midrule
\textcolor{blue}{\textbf{[Instruction]}} \\
You are a strict verifier for evaluate faithfulness. Please verify whether the given multi-table set and insight pair is faithful based on the given criteria. \\\\
\textbf{Criterion 1}: Does the insight relate to all tables within the multi-table set? \\
\textbf{Criterion 2}: Does the insight rely solely on the information within the multi-table set? \\
\textbf{Criterion 3}: Does the insight include all necessary details from the multi-table set? \\
\textbf{Criterion 4}: Does the insight provide a clear and unambiguous response based on the multi-table set? \\\\    
\textcolor{blue}{\textbf{[Inputs]}} \\
\textbf{Multi-table set}: \{serialized\_tables\} \\
\textbf{Insight}: \{insight\} \\\\
\textcolor{blue}{\textbf{[Output]}} \\
\textbf{Verification}: \\

\midrule

\textbf{Self-Verification : Completeness} \\\hline
\textcolor{blue}{\textbf{[Instruction]}} \\
You are a strict verifier for evaluate completeness. Please verify whether the given question and insight pair is complete based on the given criteria. \\\\
\textbf{Criterion 1}: Does the insight directly relate to the question? \\
\textbf{Criterion 2}: Does the insight fully address all aspects of the question? \\
\textbf{Criterion 3}: Does the insight unfold logically and without contradictions in response to the question? \\
\textbf{Criterion 4}: Does the insight remain clear and understandable in the context of the question? \\\\
\textcolor{blue}{\textbf{[Inputs]}} \\
\textbf{Question}: \{question\} \\
\textbf{Insight}: \{insight\} \\\\
\textcolor{blue}{\textbf{[Output]}} \\
\textbf{Verification}: \\

\bottomrule
\end{tabularx}
\caption{Prompts used for agent verification process.}
\label{pmt:step4_1}
\end{table*}
\begin{table*}[htbp]
\centering
\begin{tabularx}{\textwidth}{X}
\toprule

\textbf{MT-RAIG Eval - Faithfulness : Table-aware insight decomposition} \\ \midrule
\textcolor{blue}{\textbf{[Instruction]}} \\
You are a helpful assistant tasked with decomposing a given insight according to multiple table schemas. Your goal is to break the given insight down into atomic-level claims. \\\\
\textcolor{blue}{\textbf{[Inputs]}} \\
\textbf{Table schemas}: \{serialized\_table\_schemas\} \\
\textbf{Insight}: \{insight\} \\\\
\textcolor{blue}{\textbf{[Output]}} \\
\textbf{Decomposed claim set}: \\

\midrule

\textbf{MT-RAIG Eval - Faithfulness : Claim verification} \\ \midrule
\textcolor{blue}{\textbf{[Instruction]}} \\
You will be given multiple tables and a claim. Your task is to verify whether the claim is faithful to the data in the given tables.  \\\\
\textcolor{blue}{\textbf{[Inputs]}} \\
\textbf{Tables}: \{serialized\_tables\} \\
\textbf{Claim}: \{decomposed\_claim\} \\\\
\textcolor{blue}{\textbf{[Output]}} \\
\textbf{Verification}: \\
\bottomrule
\end{tabularx}
\caption{Prompts for evaluate \eval Faithfulness.}
\label{pmt:apx_faith}
\end{table*}
\begin{table*}[htbp]
\centering
\begin{tabularx}{\textwidth}{X}
\toprule

\textbf{MT-RAIG Eval - Completeness : Question-aware insight decomposition} \\ \midrule
\textcolor{blue}{\textbf{[Instruction]}} \\
You are a helpful assistant tasked with decomposing a given insight based on a question. Your goal is to extract atomic-level topics from the given insight. \\\\
\textcolor{blue}{\textbf{[Inputs]}} \\
\textbf{Question}: \{question\} \\
\textbf{Insight}: \{insight\} \\\\
\textcolor{blue}{\textbf{[Output]}} \\
\textbf{Decomposed topic set}: \\

\midrule

\textbf{MT-RAIG Eval - Completeness : Topic matching} \\ \midrule
\textcolor{blue}{\textbf{[Instruction]}} \\
You will be given two topic sets. Your task is to match topics bidirectionally based on their semantic similarity, ensuring that both sets are treated with equal importance. \\\\
\textcolor{blue}{\textbf{[Inputs]}} \\
\textbf{Topic set A}: \{decomposed\_ground\_truth\_topic\_set\} \\
\textbf{Topic set B}: \{decomposed\_predicted\_topic\_set\} \\\\
\textcolor{blue}{\textbf{[Outputs]}} \\
\textbf{Matched topic subset of A}: \\
\textbf{Matched topic subset of B}: \\
\bottomrule
\end{tabularx}
\caption{Prompts for evaluate \eval Completeness.}
\label{pmt:apx_comp}
\end{table*}
\begin{table*}[htb]
\centering
\begin{tabularx}{\textwidth}{l>{\centering\arraybackslash}X>{\centering\arraybackslash}X>{\centering\arraybackslash}X>{\centering\arraybackslash}X}
\toprule
\multirow{2.5}{*}{\textbf{Metric}} & \multicolumn{2}{c}{\textbf{Faithfulness}} & \multicolumn{2}{c}{\textbf{Completeness}} \\ \cmidrule(lr){2-3} \cmidrule(lr){4-5}
 & \textbf{Pearson} & \textbf{Spearman} & \textbf{Pearson} & \textbf{Spearman} \\ \midrule
SacreBLEU    & 31.33 & 29.11 & 33.01 & 35.80 \\
ROUGE-L      & 27.69 & 25.78 & \underline{43.43} & 41.78 \\
METEOR       & 37.00 & 36.54 & 39.96 & 37.57 \\
BERTScore    & 24.82 & 37.43 & 43.29 & \underline{42.10} \\
A3CU         & 41.22 & 41.26 & 40.10 & 39.80 \\
TAPAS-Acc    & -10.40 & -11.17 & 21.29 & 21.64 \\
G-Eval       & \underline{47.82} & \underline{46.15} & 26.35 & 23.30 \\
\rowcolor{gray!20}
\textbf{MT-RAIG Eval} & \textbf{64.94} & \textbf{65.47} & \textbf{67.67} & \textbf{68.97} \\ \midrule
Inter-Human Correlation        & 84.81 & 83.02 & 75.70 & 75.63 \\
\bottomrule
\end{tabularx}
\caption{Pearson and Spearman correlation results in terms of faithfulness and completeness.}
\label{tab:apx_meta}
\end{table*}
\begin{table*}[htbp]
\centering
\begin{tabularx}{\textwidth}{X}
\toprule

\textbf{G-Eval : Faithfulness} \\ \midrule
\textcolor{blue}{\textbf{[Instruction]}} \\
You will be given retrieved tables, a question, and an insight. Your task is to evaluate how faithful the insight is to the data in the retrieved tables when answering the question. Please make sure you read and understand these instructions carefully. \\\\
\textbf{Evaluation Criteria}: \\
Faithfulness (1–5) - A good insight must be consistent with the retrieved tables and must not include contradictory or fabricated information. \\\\
\textbf{Evaluation Steps}: \\
1. Carefully review the retrieved tables and the question, noting all critical data. \\
2. Check if the insight strictly relies on the table data without contradicting any key details. \\
3. Assign a Faithfulness score (1–5) based on the insight’s accuracy and adherence to the tables. \\\\
\textcolor{blue}{\textbf{[Inputs]}} \\
\textbf{Retrieved tables}: \{serialized\_tables\} \\
\textbf{Question}: \{question\} \\
\textbf{Insight}: \{insight\} \\\\
\textcolor{blue}{\textbf{[Output]}} \\
\textbf{Faithfulness (scores ONLY)}: \\

\midrule

\textbf{G-Eval : Completeness} \\ \midrule
\textcolor{blue}{\textbf{[Instruction]}} \\
You will be given retrieved tables, a question, and an insight. Your task is to evaluate whether the insight fully addresses the question using the data provided. Please make sure you read and understand these instructions carefully. \\\\
\textbf{Evaluation Criteria}: \\
Completeness (1–5) - A good insight must address all relevant elements of the question, leaving no key points from the tables unaddressed. \\\\
\textbf{Evaluation Steps}: \\
1. Review the retrieved tables and question to identify the essential information. \\
2. Compare the insight against the question and tables, checking if all crucial points are covered. \\
3. Assign a Completeness score (1–5) based on how thoroughly the insight uses the table data to fulfill the question. \\\\
\textcolor{blue}{\textbf{[Inputs]}} \\
\textbf{Retrieved tables}: \{serialized\_tables\} \\
\textbf{Question}: \{question\} \\
\textbf{Insight}: \{insight\} \\\\
\textcolor{blue}{\textbf{[Output]}} \\
\textbf{Completeness (scores ONLY)}: \\
\bottomrule
\end{tabularx}
\caption{Prompts used for G-Eval.}
\label{pmt:apx_geval}
\end{table*}
\begin{table*}[htb]
\centering
\begin{tabularx}{\textwidth}{l *{12}{>{\centering\arraybackslash}X}}
\toprule
\multirow{2.5}{*}{\textbf{Retriever}} & \multicolumn{3}{c}{\textbf{Top-2}} & \multicolumn{3}{c}{\textbf{Top-5}} & \multicolumn{3}{c}{\textbf{Top-10}} & \multicolumn{3}{c}{\textbf{Top-20}} \\
\cmidrule(lr){2-4} \cmidrule(lr){5-7} \cmidrule(lr){8-10} \cmidrule(lr){11-13}
 & \textbf{P} & \textbf{R} & \textbf{F1} & \textbf{P} & \textbf{R} & \textbf{F1} & \textbf{P} & \textbf{R} & \textbf{F1} & \textbf{P} & \textbf{R} & \textbf{F1} \\
\midrule
\rowcolor{gray!20}
\multicolumn{13}{l}{\textit{General-purpose retrievers}} \\
BM25
& 21.07 & 17.26 & 18.98
& 14.06 & 27.19 & 18.54
&  8.85 & 33.72 & 14.02
&  5.46 & 41.16 &  9.65 \\
DPR
& \textbf{58.77} & \textbf{44.58} & \textbf{50.70}
& \textbf{37.92} & \textbf{68.24} & \textbf{48.75}
& \textbf{22.61} & \textbf{80.83} & \textbf{35.26}
& \textbf{12.62} & \textbf{88.45} & \textbf{22.08} \\
Contriever
& 29.69 & 23.47 & 26.22
& 19.14 & 35.67 & 24.91
& 12.19 & 44.27 & 19.11
&  7.31 & 52.12 & 12.82 \\
\rowcolor{gray!20}
\multicolumn{13}{l}{\textit{Table-specific retrievers}} \\
DTR
& \underline{49.40} & \underline{37.77} & \underline{42.81}
& \underline{32.88} & \underline{59.60} & \underline{42.38}
& \underline{21.04} & \underline{74.50} & \underline{32.81}
& \underline{12.37} & \underline{86.22} & \underline{21.63} \\
TableLlama
& 47.99 & 36.93 & 41.74
& 32.70 & 59.48 & 42.20
& 20.34 & 72.44 & 31.76
& 11.55 & 81.56 & 20.24 \\
\bottomrule
\end{tabularx}
\caption{Multi-table retrieval results in terms of P/R/F1.}
\label{tab:apx_rtr}
\end{table*}
\begin{table*}[htb]
\centering
\begin{scriptsize}
\begin{tabularx}{\textwidth}{l*{12}{>{\centering\arraybackslash}X}||*{12}{>{\centering\arraybackslash}X}}
\toprule
\multirow{4}{*}{\textbf{Generator}} & \multicolumn{15}{c}{\textbf{Completeness}} \\ \cmidrule(lr){2-16}
& \multicolumn{3}{c}{A\&S} & \multicolumn{3}{c}{C\&R} & \multicolumn{3}{c}{P\&O} & \multicolumn{3}{c}{T\&P} & \multicolumn{3}{c}{Avg.} \\ \cmidrule(lr){2-4} \cmidrule(lr){5-7} \cmidrule(lr){8-10} \cmidrule(lr){11-13} \cmidrule(lr){14-16}
& \textbf{P} & \textbf{R} & \textbf{F1} & \textbf{P} & \textbf{R} & \textbf{F1} & \textbf{P} & \textbf{R} & \textbf{F1} & \textbf{P} & \textbf{R} & \textbf{F1} & \textbf{P} & \textbf{R} & \textbf{F1} \\ \midrule
\rowcolor{gray!20}
\multicolumn{16}{l}{\textit{Proprietary LLMs}} \\ 
\includegraphics[width=0.8em]{sourceEMOJI/OpenAI.png} o3-mini
& 67.40 & 62.92 & 62.84
& \underline{63.78} & 57.55 & \underline{58.78}
& \underline{65.65} & 57.69 & 59.77
& \textbf{65.28} & 57.33 & 59.42
& \underline{65.53} & 58.87 & 60.20 \\
\includegraphics[width=0.8em]{sourceEMOJI/OpenAI.png} GPT-4o
& 66.03 & \textbf{66.68} & \textbf{64.46}
& 61.93 & \textbf{60.06} & \textbf{59.37}
& 64.65 & \textbf{61.54} & \textbf{61.49}
& 63.22 & \textbf{61.55} & \textbf{60.77}
& 63.96 & \textbf{62.46} & \textbf{61.52} \\
\includegraphics[width=0.8em]{sourceEMOJI/Claude.png} Claude 3.5 Sonnet
& 62.20 & \underline{64.67} & 61.67
& 56.45 & 58.56 & 56.06
& 60.78 & 60.21 & 58.94
& 59.02 & \underline{59.54} & 57.86
& 59.61 & \underline{60.75} & 58.63 \\
\midrule
\rowcolor{gray!20}
\multicolumn{16}{l}{\textit{Open-source LLMs}} \\
\includegraphics[width=0.8em]{sourceEMOJI/DeepSeek.png} DeepSeek-R1-8B
& \underline{68.61} & 64.24 & \underline{64.23}
& 62.21 & 58.15 & 58.72
& 65.29 & \underline{60.89} & \underline{61.41}
& 62.92 & 59.02 & 59.48
& 64.76 & 60.58 & \underline{60.96}  \\
\includegraphics[width=0.8em]{sourceEMOJI/Qwen.png} Qwen2-7B
& 64.62 & 63.25 & 61.92
& 60.61 & \underline{58.31} & 57.88
& 62.79 & 59.83 & 59.73
& 59.87 & 57.99 & 57.32
& 61.97 & 59.84 & 59.21 \\
\includegraphics[width=0.8em]{sourceEMOJI/Google.png} Gemma-7B
& \textbf{68.76} & 61.31 & 62.72
& \textbf{65.93} & 54.43 & 57.73
& \textbf{67.06} & 58.29 & 60.52
& \underline{65.28} & 57.80 & \underline{59.68}
& \textbf{66.76} & 57.96 & 60.16 \\
\includegraphics[width=0.8em]{sourceEMOJI/Meta.png} Llama 3.1-8B
& 59.24 & 63.43 & 59.20
& 56.21 & 57.16 & 55.11
& 58.95 & 59.44 & 57.45
& 57.02 & 58.07 & 55.85
& 57.85 & 59.52 & 56.90 \\
\includegraphics[width=0.8em]{sourceEMOJI/Mistral.png} Mistral-7B
& 66.62 & 63.69 & 62.96
& 62.16 & 57.26 & 58.00
& 63.53 & 59.13 & 59.66
& 62.05 & 57.80 & 58.29
& 63.59 & 59.47 & 59.73 \\
\midrule
\rowcolor{gray!20}
\multicolumn{16}{l}{\textit{SOTA TQA-methods}} \\
\includegraphics[width=0.8em]{sourceEMOJI/Damo.png} Dater
& 63.53 & 59.97 & 59.70
& 60.57 & 56.96 & 57.26
& 64.53 & 59.23 & 60.35
& 62.20 & 59.12 & 59.16
& 62.71 & 58.82 & 59.12 \\
\includegraphics[width=0.8em]{sourceEMOJI/Google.png} Chain-of-Table
& 62.26 & 59.84 & 59.29
& 59.34 & 54.78 & 55.49
& 62.02 & 58.08 & 58.58
& 60.22 & 57.12 & 57.17
& 60.96 & 57.46 & 57.63 \\
\includegraphics[width=0.8em]{sourceEMOJI/Yale.png} TaPERA
& 59.82 & 62.51 & 59.23
& 53.69 & 57.31 & 53.56
& 60.25 & 59.32 & 58.05
& 55.35 & 59.12 & 55.53
& 57.28 & 59.57 & 56.59 \\
\bottomrule
\end{tabularx}
\end{scriptsize}
\caption{\eval Completeness score results in terms of P/R/F1.}
\label{tab:apx_comp}
\end{table*}
\begin{table*}[htb]
\centering
\begin{scriptsize}
\begin{tabularx}{\textwidth}{l>{\centering\arraybackslash}X>{\centering\arraybackslash}X>{\centering\arraybackslash}X>{\centering\arraybackslash}X>{\centering\arraybackslash}X>{\centering\arraybackslash}X>{\centering\arraybackslash}X>{\centering\arraybackslash}X}
\toprule
\multirow{2.5}{*}{\textbf{Generator}} & \multirow{2.5}{*}{\textbf{SacreBLEU}} & \multirow{2.5}{*}{\textbf{ROUGE-L}} & \multirow{2.5}{*}{\textbf{METEOR}} & \multirow{2.5}{*}{\textbf{BERTScore}} & \multirow{2.5}{*}{\textbf{A3CU}} & \multirow{2.5}{*}{\textbf{TAPAS-Acc}} & \multicolumn{2}{c}{\textbf{G-Eval}} \\ \cmidrule(lr){8-9}
 & & & & & & & \textbf{Faithfulness} & \textbf{Completeness} \\ \midrule
\rowcolor{gray!20}
\multicolumn{9}{l}{\textit{Proprietary LLMs}} \\
\includegraphics[width=0.8em]{sourceEMOJI/OpenAI.png} o3-mini
&  7.75 & 24.61 & 29.12
& 87.94 & 25.63 & 39.65
& 4.25 & \underline{4.64} \\
\includegraphics[width=0.8em]{sourceEMOJI/OpenAI.png} GPT-4o
& \textbf{19.16} & \textbf{33.92} & \textbf{38.91}
& \textbf{90.02} & \textbf{35.87} & 34.15
& \textbf{4.34} & \textbf{4.69} \\
\includegraphics[width=0.8em]{sourceEMOJI/Claude.png} Claude 3.5 Sonnet
& 15.59 & 30.11 & 35.99
& \underline{89.40} & 31.87 & 31.77
& \underline{4.28} & 4.51 \\
\midrule
\rowcolor{gray!20}
\multicolumn{9}{l}{\textit{Open-source LLMs}} \\
\includegraphics[width=0.8em]{sourceEMOJI/DeepSeek.png} DeepSeek-R1-8B
& 12.07 & 30.03 & 28.48
& 89.22 & 31.32 & 54.70
& 3.85 & 4.25 \\
\includegraphics[width=0.8em]{sourceEMOJI/Qwen.png} Qwen2-7B
& 17.20 & 32.06 & \underline{36.77}
& 89.43 & 30.41 & 38.23
& 3.56 & 4.32 \\
\includegraphics[width=0.8em]{sourceEMOJI/Google.png} Gemma-7B
& 12.31 & 28.83 & 26.96
& 88.41 & 23.91 & \textbf{64.05}
& 3.05 & 3.61 \\
\includegraphics[width=0.8em]{sourceEMOJI/Meta.png} Llama 3.1-8B
&  9.55 & 23.85 & 35.62
& 86.45 & 27.03 & \underline{61.25}
& 3.57 & 4.14 \\
\includegraphics[width=0.8em]{sourceEMOJI/Mistral.png} Mistral-7B
& 15.07 & 30.20 & 34.66
& 89.00 & 27.51 & 43.12
& 3.55 & 4.30 \\
\midrule
\rowcolor{gray!20}
\multicolumn{9}{l}{\textit{SOTA TQA-methods}} \\
\includegraphics[width=0.8em]{sourceEMOJI/Damo.png} Dater
& 18.06 & \underline{32.50} & 34.94
& 89.39 & 29.40 & 43.00
& 3.28 & 4.01 \\
\includegraphics[width=0.8em]{sourceEMOJI/Google.png} Chain-of-Table
& \underline{18.15} & 32.38 & 35.60
& 87.33 & \underline{33.26} & 42.08
& 3.69 & 4.25 \\
\includegraphics[width=0.8em]{sourceEMOJI/Yale.png} TaPERA
& 13.89 & 28.57 & 32.74
& 88.32 & 24.61 & 44.67
& 2.63 & 4.00 \\
\bottomrule
\end{tabularx}
\end{scriptsize}
\caption{Insight generation results alongside existing automatic metrics. We report the average scores across all question types.}
\label{tab:apx_autoeval}
\end{table*}
\begin{table*}[htbp]
\centering
\begin{small}
\begin{tabularx}{\textwidth}{%
>{\arraybackslash}p{0.15\textwidth}
>{\arraybackslash}X
}
\toprule
\multicolumn{2}{l}{\textbf{Question type : Analysis \& Summary}} \\ \midrule
\textbf{Multi-table set}
& \textbf{Table 1} \\
& [TITLE] Spain in the Eurovision Song Contest 2009 - Semi-final 1 - Semi-final 1 - 14 February 2009 \\
& [HEADER] Draw | Artist | Song | Jury\_votes | Televotes | Total\_votes | Result\_ \\
& [ROW 1] 1 | Yulia Valentayn | Uh la la | 6 | 3 | 9 | Out \\
& [ROW 2] 2 | La Red de San Luis | Gracias, Madre Tierra | 8 [b ] | 2 | 10 | Out \\
& [ROW 3] 3 | Vicente Casal | Tú me complementas | 1 | 7 | 8 | Out \\
& ... \\
& \textbf{Table 2} \\
& [TITLE] Spain in the Eurovision Song Contest 2009 - Semi-final 2 - Semi-final 2 - 21 February 2009 \\
& [HEADER] Draw | Artist | Song | Jury\_votes | Televotes | Total\_votes | Result\_ \\
& [ROW 1] 1 | Diqesi | Subiré | 5 | 4 | 9 | Out \\
& [ROW 2] 2 | Roel | Y ahora dices | 6 | 3 | 9 | Out \\
& [ROW 3] 3 | Salva Ortega | Lujuria | 7 | 7 | 14 | Second Chance > Final \\
& ... \\
& \textbf{Table 3} \\
& [TITLE] Spain in the Eurovision Song Contest 2009 - Semi-final 3 - Semi-final 3 - 28 February 2009 \\
& [HEADER] Draw | Artist | Song | Jury\_votes | Televotes | Total\_votes | Result\_ \\
& [ROW 1] 1 | El Secreto de Álex | Por esta vez | 7 | 6 | 13 | Out \\
& [ROW 2] 2 | Beatriz | ¿Dónde te metes? | 4 | 1 | 5 | Out \\
& [ROW 3] 3 | Remembrances | Espejismo | 3 | 4 | 7 | Out \\
& ... \\ \midrule
\textbf{Question}
& Summarize the performances of the artists who qualified for the final from each of the three semi-finals in Spain's Eurovision Song Contest 2009. What were their total votes, and how did they compare to those who did not qualify? \\ \midrule
\textbf{Insight}
& In Spain's Eurovision Song Contest 2009, the artists who qualified for the final from the three semi-finals showcased varying levels of performance based on their total votes. In the first semi-final, Noelia Cano received 11 votes, La La Love You garnered 20 votes, and Melody y Los Vivancos achieved 24 votes. Comparatively, those who did not qualify, such as Yulia Valentayn and La Red de San Luis, received 9 and 10 votes respectively, indicating that the qualifiers had a stronger reception. In the second semi-final, Soraya led with 24 votes, followed closely by Virginia with 20 votes and Jorge González with 16 votes, while non-qualifiers like Diqesi and Roel only managed 9 votes each, highlighting a clear distinction in popularity. The third semi-final saw Mirela, Isi, Santa Fe, and Julia Bermejo qualify with 18, 17, 24, and 14 votes respectively, while non-qualifiers like El Secreto de Álex and Beatriz received 13 and 5 votes, further emphasizing the gap in support. Overall, the artists who qualified consistently received higher votes than those who did not, reflecting their stronger performances and audience appeal across all semi-finals. \\ \midrule
\multicolumn{2}{l}{\textbf{Question type : Comparison \& Relationship}} \\ \midrule
\textbf{Multi-table set}
& \textbf{Table 1} \\
& [TITLE] car\_1 - car\_makers \\
& [HEADER] Id | Maker | FullName | Country \\
& [ROW 1] 1 | amc | American Motor Company | 1 \\
& [ROW 2] 2 | volkswagen | Volkswagen | 2 \\
& [ROW 3] 3 | bmw | BMW | 2 \\
& ... \\
& \textbf{Table 2} \\
& [TITLE] car\_1 - model\_list \\
& [HEADER] ModelId | Maker | Model \\
& [ROW 1] 1 | 1 | amc \\
& [ROW 2] 2 | 2 | audi \\
& [ROW 3] 3 | 3 | bmw \\
& ... \\
\midrule
\textbf{Question} &
How does the id of a car maker relate to the specific models in the model list, and what does this indicate about the organization of car manufacturing data? \\ \midrule
\textbf{Insight} &
The id of a car maker is directly linked to the specific models in the model list, as each model is associated with a unique maker id that identifies the manufacturer responsible for it. For instance, the id 13 corresponds to Daimler Benz, which produces models like Mercedes-Benz and Mercedes, while id 19 is associated with Toyota, which includes models such as Scion and Toyota. This structured relationship indicates that car manufacturing data is organized in a way that allows for easy identification and categorization of models under their respective makers, facilitating data management and retrieval. Furthermore, the presence of makers like General Motors and Chrysler, which have designed more than three models, highlights how certain manufacturers have a broader portfolio, thus enriching the dataset and providing insights into the diversity of offerings from each maker. Overall, this organization enhances clarity and efficiency in understanding the connections between car makers and their models. \\
\bottomrule
\end{tabularx}
\end{small}
\caption{\bench example for two question types, Analysis \& Summary and Comparison \& Relationship.}
\label{case:bench_i}
\end{table*}
\begin{table*}[htbp]
\centering
\begin{small}
\begin{tabularx}{\textwidth}{%
>{\arraybackslash}p{0.15\textwidth}
>{\arraybackslash}X
}
\toprule
\multicolumn{2}{l}{\textbf{Question type : Trend \& Pattern}} \\ \midrule
\textbf{Multi-table set}
& \textbf{Table 1} \\
& [TITLE] La Cage aux Folles (musical) - 2004 Broadway revival - 2004 Broadway revival \\
& [HEADER] Year\_ | Award | Category | Nominee | Result\_ \\
& [ROW 1] 2005 | Tony Award | Best Revival of a Musical | Best Revival of a Musical | Won \\
& [ROW 2] 2005 | Tony Award | Best Performance by a Leading Actor in a Musical | Gary Beach | Nominated \\
& [ROW 3] 2005 | Tony Award | Best Choreography | Jerry Mitchell | Won \\
& \textbf{Table 2} \\
& [TITLE] La Cage aux Folles (musical) - 2010 Broadway revival - 2010 Broadway revival \\
& [HEADER] Year\_ | Award | Category | Nominee | Result\_ \\
& [ROW 1] 2010 | Tony Award | Best Revival of a Musical | Best Revival of a Musical | Won \\
& [ROW 2] 2010 | Tony Award | Best Performance by a Leading Actor in a Musical | Kelsey Grammer | Nominated \\
& [ROW 3] 2010 | Tony Award | Best Performance by a Leading Actor in a Musical | Douglas Hodge | Won \\ \midrule
\textbf{Question}
& Compare the categories in which La Cage aux Folles was nominated for the Best Revival of a Musical in both the 2004 and 2010 Broadway revivals. What trends can be observed in the types of performances and productions recognized in these years? \\ \midrule
\textbf{Insight}
& In comparing the nominations for Best Revival of a Musical for La Cage aux Folles in the 2004 and 2010 Broadway revivals, several trends emerge regarding the types of performances and productions recognized. In 2004, the revival won the Tony Award for Best Revival and received nominations primarily focused on choreography, with Jerry Mitchell winning for Best Choreography, and a nomination for Best Performance by a Leading Actor in a Musical for Gary Beach. The emphasis was notably on choreography and costume design, reflecting a strong appreciation for the visual and performative aspects of the production. In contrast, the 2010 revival also won the Tony Award for Best Revival but expanded its recognition to include multiple categories, highlighting a broader range of performances with nominations for both Kelsey Grammer and Douglas Hodge in the Leading Actor category, as well as Robin de Jesus in the Featured Actor category. Additionally, the 2010 revival received nominations for Best Direction and various technical design categories, indicating a trend towards recognizing not only performances but also the overall production quality and direction. This shift suggests an evolving appreciation for comprehensive theatrical artistry, where both individual performances and the collaborative efforts of direction and design are increasingly acknowledged. \\ \midrule
\multicolumn{2}{l}{\textbf{Question type : Performance \& Outcome}} \\ \midrule
\textbf{Multi-table set}
& \textbf{Table 1} \\
& [TITLE] Elissa Sursara - Filmography - Filmography \\
& [HEADER] Year\_ | Film | Role\_ | Language\_ | Notes \\
& [ROW 1] 2000 | Hector | Reno | English | Film \\
& [ROW 2] 2002 | Marion Bridge | Sam | English | Film \\
& [ROW 3] 2002 | The Wet Season | Taylor | English | Film \\
& ... \\
& \textbf{Table 2} \\
& [TITLE] Elissa Sursara - Television - Television \\
& [HEADER] Year\_ | Title | Role\_ | Language\_ | Notes \\
& [ROW 1] 1994-96 | Para Crancias | Co-Host | Portuguese | 22 episodes \\
& [ROW 2] 1994-96 | Touched by an Angel | Natalie | English | The Heart of the Matter \\
& [ROW 3] 1996 | American Gothic | Alex | English | Requiem \\
& ... \\ \midrule
\textbf{Question}
& What roles did Elissa Sursara undertake in the years 2002 and 2007 across her filmography and television appearances, and how do these roles reflect her career progression during that period? \\ \midrule
\textbf{Insight}
& In 2002, Elissa Sursara took on the roles of Sam in "Marion Bridge" and Taylor in "The Wet Season," alongside portraying Lena in the TV show "Taken," which positioned her primarily in supporting roles. By 2007, her filmography expanded significantly as she played Ava in "The Screening," Bena Brazilian in "Equal Opportunity," and Brittany in "Pain Within," while also taking on the character Jessica in the television series "Brothers \& Sisters." This evolution from supporting roles in 2002 to more varied and prominent characters in 2007 reflects her career progression, highlighting her growing versatility and ability to engage with diverse genres and complex characters, marking a significant development in her acting career during that period. \\
\bottomrule
\end{tabularx}
\end{small}
\caption{\bench example for two question types, Trend \& Pattern and Performance \& Outcome.}
\label{case:bench_ii}
\end{table*}
\begin{table*}[htbp]
\centering
\begin{small}
\begin{tabularx}{\textwidth}{%
>{\arraybackslash}p{0.15\textwidth}
>{\arraybackslash}X
}
\toprule
\multicolumn{2}{l}{\textbf{Criteria : Relevance}} \\ \midrule
\textbf{Multi-table set}
& \textbf{Table 1} \\
& [TITLE] European Junior Swimming Championships | Men | Men \\
& [HEADER] Event | Time\_ | Nationality | Date\_ | Meet | Location\_ \\
& [ROW 1] 50 m freestyle | 22.06 | Russia | 14 July 2013 | 2013 Championships | Poznan , Poland \\
& [ROW 2] 100 m freestyle | 48.48 | Russia | 10 July 2009 | 2009 Championships | Prague , Czech Republic \\
& [ROW 3] 200 m freestyle | 1:46.58 | France | 18 July 2010 | 2010 Championships | Helsinki , Finland \\
& ... \\
& \textbf{Table 2} \\
& [TITLE] FINA World Junior Swimming Championships | Men | Men \\
& [HEADER] Event | Time\_ | Nationality | Date\_ | Meet | Location\_ \\ \\
& [ROW 1] 50m freestyle | 22.11 | Australia | 28 August 2013 | 2013 Championships | Dubai , United Arab Emirates \\
& [ROW 2] 100m freestyle | 48.97 | United States | 31 August 2013 | 2013 Championships | Dubai , United Arab Emirates \\
& [ROW 3] 200m freestyle | 1:47.55 | Australia | 27 August 2013 | 2013 Championships | Dubai , United Arab Emirates \\
& ... \\
& \textbf{Table 3} \\
& [TITLE] List of Universiade records in swimming | Men | Men \\
& [HEADER] Event | Time\_ | Nationality | Date\_ | Meet | Location\_ \\
& [ROW 1] 50m freestyle | 21.67 | Russia | 16 July 2013 | 2013 Universiade | Kazan , Russia \\
& [ROW 2] 100m freestyle | 47.62 | Russia | 14 July 2013 | 2013 Universiade | Kazan , Russia \\
& [ROW 3] 200m freestyle | 1:44.87 | Russia | 15 July 2013 | 2013 Universiade | Kazan , Russia \\
& ... \\ \midrule
\textbf{Question}
& Analyze the significance of the locations where the European Junior Swimming Championships and the Universiade records in swimming are held. How do these venues contribute to the overall experience and performance of the athletes? \\ \midrule
\textbf{Verification}
& The question does not relate to all the tables in the set, as it primarily mentions the European Junior \\
\textbf{rationale}
& Swimming Championships and Universiade records, while the other tables (such as the Hungarian records) are not considered. \\ \midrule
\multicolumn{2}{l}{\textbf{Criteria : Faithfulness}} \\ \midrule
\textbf{Multi-table set}
& \textbf{Table 1} \\
& [TITLE] Aqeel Khan | Singles runner-up | Singles runner-up \\
& [HEADER] Date\_ | Tournament | Surface | Opponent\_in\_the\_final | Score \\
& [ROW 1] 28 June 2004 | Gurgaon | Hard | Norikazu Sugiyama | 6-1 3-6 4-6 \\
& [ROW 2] 16 August 2004 | Lahore | Grass | Toshiaki Sakai | 4-6 6-7 \\
& [ROW 3] 17 August 2006 | Delhi | Hard | Yuri Bezeruk | 6-7 2-6 \\
& ... \\
& \textbf{Table 2} \\
& [TITLE] Aqeel Khan | Singles titles | Singles titles \\
& [HEADER] Date\_ | Tournament | Surface | Opponent\_in\_the\_final | Score \\
& [ROW 1] 15 August 2004 | Islamabad | Clay | Toshiaki Sakai | 7-6(3) 7-6(5) \\
& [ROW 2] 22 August 2004 | Lahore | Grass | Toshiaki Sakai | 1-6 6-4 6-3 \\
& [ROW 3] 5 September 2004 | Karachi | Hard | Tommaso Sanna | 6-3 6-4 \\
& ... \\ \midrule
\textbf{Insight}
& Aqeel Khan faced Toshiaki Sakai in two finals, first on 15 August 2004 at the Islamabad tournament on Clay, where he won with a score of 7-6(3) 7-6(5), demonstrating strong performance and resilience. He then met Sakai again on 22 August 2004 at the Lahore tournament on Grass, where he lost 1-6 6-4 6-3, indicating his ability to recover after a poor start. Additionally, Khan faced Divij Sharan on 29 October 2007 at the Lahore tournament, also on Grass, where he lost 4-6 6-3 6-4, showcasing his competitive spirit as he managed to win the second set after losing the first. These matchups highlight Khan's competitive nature and adaptability, as he demonstrated the ability to perform well under pressure and recover from setbacks, even though he did not secure victories in all his finals. \\ \midrule
\textbf{Verification}
& The answer does not cover all tables, focusing only on Table 2 (Aqeel Khan’s singles titles) while \\
\textbf{rationale}
& omitting matches from Table 1 (Aqeel Khan’s singles runner-up results). It also fails to fully integrate data from both tables. \\
\bottomrule
\end{tabularx}
\end{small}
\caption{Discarded data example in self-verification process for criteria about relevance and faithfulness.}
\label{case:filter_i}
\end{table*}
\begin{table*}[htbp]
\centering
\begin{small}
\begin{tabularx}{\textwidth}{%
>{\arraybackslash}p{0.15\textwidth}
>{\arraybackslash}X
}
\toprule
\multicolumn{2}{l}{\textbf{Criteria : Completeness}} \\ \midrule
\textbf{Question}
& Analyze the character arcs presented in the episodes directed by James Ormerod in Series 2: Spearhead in Germany and Series 3: Spearhead in Hong Kong. How do these arcs contribute to the overarching narrative of the Spearhead (TV series)? \\ \midrule
\textbf{Insight}
& The episodes directed by James Ormerod in Series 2 and Series 3 of Spearhead, specifically "New Brooms," "Matrimony," "Pressures," "Leave Takers," and "Repercussions" from Series 2, along with "Have a Happy Day" and "Another Happy Day" from Series 3, present character arcs that are integral to the series' overarching narrative, although specific details about these arcs are not provided in the available knowledge. The character development across these episodes likely reflects themes of personal growth, conflict resolution, and the impact of relationships, which are common in serialized storytelling. By exploring the dynamics between characters and their responses to various pressures and life changes, these arcs contribute to a deeper understanding of the characters' motivations and the series' exploration of societal issues. However, without explicit details on the character arcs, one can infer that the evolution of characters in these episodes serves to enhance the narrative by illustrating the complexities of human experiences in the context of the show's broader themes. \\ \midrule
\textbf{Verification}
& The answer does not fully address the question, as it lacks specific details about the character arcs in \\
\textbf{rationale}
& the episodes directed by James Ormerod. While it touches on general themes and character development, it does not analyze the specific arcs mentioned, which are crucial for a complete response. \\
\bottomrule
\end{tabularx}
\end{small}
\caption{Discarded data example in self-verification process for criteria about completeness.}
\label{case:filter_ii}
\end{table*}
\begin{table*}[htbp]
\centering
\begin{small}
\begin{tabularx}{\textwidth}{%
>{\arraybackslash}p{0.15\textwidth}
>{\arraybackslash}X
}
\toprule
\multicolumn{2}{l}{\textbf{\eval Faithfulness score :} 47.37} \\ \midrule
\multicolumn{2}{l}{\textbf{Step : Table-aware insight decomposition}} \\ \midrule
\textbf{Table schemas}
& \textbf{Table 1} \\
& [TITLE] 1988 Notre Dame Fighting Irish football team - Schedule - Schedule \\
& [HEADER] Date\_ | Time\_ | Opponent\_ | Rank\_ | Result\_ | Attendance \\
& \textbf{Table 2} \\
& [TITLE] 2005 Notre Dame Fighting Irish football team - Schedule - Schedule \\
& [HEADER] Date\_ | Time\_ | Opponent\_ | Rank\_ | Site | TV | Result\_ | Attendance\_ \\
& \textbf{Table 3} \\
& [TITLE] 1946 Notre Dame Fighting Irish football team - Schedule - Schedule \\
& [HEADER] Date\_ | Opponent\_ | Rank\_ | Site | Result\_ \\
& ... \\
& \textbf{Table 10} \\
& [TITLE] 1988 Minnesota Vikings season - Schedule - Schedule \\
& [HEADER] Week | Date\_ | Opponent | Result\_ | Attendance \\
\midrule
\textbf{Insight}
& The 1946 and 1988 Notre Dame Fighting Irish football teams both demonstrated impressive performances, but with notable differences in their ranking trajectories and strength of schedule. In 1946, Notre Dame started unranked but quickly ascended to \#3 by their third game and maintained a \#2 ranking for the majority of the season, facing only two ranked opponents. The 1988 team, however, began the season ranked \#13 and steadily climbed to \#1 by their ninth game, maintaining that top position for the remainder of the season. The 1988 schedule appears more challenging, with five ranked opponents, including a \#1 Miami team, compared to the 1946 schedule. This suggests that the 1988 team faced a tougher slate of opponents, yet still managed to achieve and maintain the top ranking. The 1988 team's ability to consistently improve their ranking against stronger competition may indicate a more impressive overall season performance compared to the 1946 team, which maintained a high but steady ranking against fewer ranked opponents. \\ \midrule
\textbf{Decomposed}
& 1. In 1946, Notre Dame started unranked but quickly ascended to \#3 by their third game. \\
\textbf{claim set}
& 2. In 1946, Notre Dame maintained a \#2 ranking for the majority of the season. \\
& 3. In 1946, Notre Dame faced only two ranked opponents. \\
& 4. The 1988 team began the season ranked \#13. \\
& 5. The 1988 team steadily climbed to \#1 by their ninth game. \\
& 6. The 1988 team maintained the top position for the remainder of the season. \\
& 7. The 1988 schedule included five ranked opponents. \\
& 8. The 1988 schedule included a \#1 Miami team. \\
& 9. The 1946 schedule had fewer ranked opponents compared to the 1988 schedule. \\ \midrule
\multicolumn{2}{l}{\textbf{Step : Claim verification}} \\ \midrule
\textbf{Tables}
& \textit{As referenced above} \\ \midrule
\textbf{Claim}
& In 1946, Notre Dame maintained a \#2 ranking for the majority of the season. \\ \midrule
\textbf{Verification}
& [1.0, 1.0, 1.0, 1.0, 0.0, 1.0, 0.0, 0.0, 1.0, 1.0, 1.0, 0.0, 1.0, 1.0, 0.0, 0.0, 1.0, 1.0, 0.0, 1.0] \\
\bottomrule
\end{tabularx}
\end{small}
\caption{\eval Faithfulness evaluation example for each step based on Claude 3.5 Sonnet.}
\label{case:eval_faith}
\end{table*}
\begin{table*}[htbp]
\centering
\begin{small}
\begin{tabularx}{\textwidth}{%
>{\arraybackslash}p{0.15\textwidth}
>{\arraybackslash}X
}
\toprule
\multicolumn{2}{l}{\textbf{\eval Completeness score :} 69.57} \\ \midrule
\multicolumn{2}{l}{\textbf{Step : Question-aware insight decomposition}} \\ \midrule
\textbf{Question}
& Summarize the attendance figures for the 1968 Buffalo Bills season and the 1969 Buffalo Bills season, focusing on the games against the Houston Oilers and Oakland Raiders. What might these figures indicate about fan engagement and support for the team during these years? \\ \midrule
\textbf{Ground-truth}
& The attendance figures for the 1968 Buffalo Bills season against the Houston Oilers were 34,339 on \\
\textbf{insight}
& October 27 and 34,110 on December 7, while the games against the Oakland Raiders drew 43,056 on September 15 and 39,883 on November 28, indicating a moderate level of fan engagement despite the team's struggles that season. In contrast, the 1969 season saw increased attendance figures against the Houston Oilers, with 40,146 on September 21 and 46,485 on October 5, and a significant turnout of 54,418 for the game against the Oakland Raiders on October 19. These rising attendance numbers from 1968 to 1969 suggest a growing support and enthusiasm for the team, reflecting an improvement in fan engagement as the Bills likely began to show better performance or potential, which could have contributed to increased interest and attendance at their games. \\ \midrule
\textbf{Decomposed}
& 1. Attendance figures for 1968 Buffalo Bills season. \\
\textbf{ground-truth}
& 2. Attendance figures for 1969 Buffalo Bills season. \\
\textbf{topic set}
& 3. Attendance figures against Houston Oilers in 1968. \\
& 4. Attendance figures against Oakland Raiders in 1968. \\
& 5. Attendance figures against Houston Oilers in 1969. \\
& 6. Attendance figures against Oakland Raiders in 1969. \\
& 7. Moderate level of fan engagement in 1968. \\
& 8. Increased attendance figures in 1969. \\
& 9. Growing support for the team. \\
& 10. Improvement in fan engagement. \\
& 11. Team performance impact on attendance. \\
& 12. Enthusiasm for the team. \\ \midrule
\textbf{Predicted}
& In the 1968 Buffalo Bills season, the home game against the Houston Oilers on October 27 drew an \\
\textbf{insight}
& attendance of 34,339, while the away game against the Oakland Raiders on November 28 had 39,883 attendees. In the 1969 season, the home game against the Houston Oilers on September 21 saw a slightly higher attendance of 40,146, and the away game against the Oakland Raiders on October 19 attracted 54,418 fans. These figures suggest that despite the Bills' poor performance in both seasons, with records of 1-12-1 in 1968 and 4-10 in 1969, there was a modest increase in attendance for these matchups, particularly in 1969. The higher attendance for the away games, especially against the Raiders, may reflect stronger fan engagement in cities with more competitive teams, while the relatively lower home attendance could indicate waning local support due to the Bills' struggles on the field. This trend highlights how team performance and the competitiveness of opponents can influence fan turnout and engagement. \\ \midrule
\textbf{Decomposed}
& 1. Attendance figures for 1968 Buffalo Bills season. \\
\textbf{predicted}
& 2. Attendance figures for 1969 Buffalo Bills season. \\
\textbf{topic set}
& 3. Attendance at home game against Houston Oilers. \\
& 4. Attendance at away game against Oakland Raiders. \\
& 5. Comparison of attendance figures between seasons. \\
& 6. Modest increase in attendance for matchups. \\
& 7. Higher attendance for away games. \\
& 8. Fan engagement in cities with competitive teams. \\
& 9. Lower home attendance for Buffalo Bills. \\
& 10. Impact of team performance on local support. \\
& 11. Influence of opponent competitiveness on fan turnout. \\ \midrule
\multicolumn{2}{l}{\textbf{Step : Topic matching}} \\ \midrule
\textbf{Ground-truth}
& \textit{As referenced above} \\
\textbf{topic set} & \\ \midrule
\textbf{Predicted}
& \textit{As referenced above} \\
\textbf{topic set} & \\ \midrule
\textbf{Matched}
& [1, 2, 3, 4, 5, 6, 8, 11] \\
\textbf{ground-truth} & \\
\textbf{topic subset} & \\ \midrule
\textbf{Matched}
& [1, 2, 3, 4, 5, 6, 10, 11] \\
\textbf{predicted} & \\
\textbf{topic subset} & \\
\bottomrule
\end{tabularx}
\end{small}
\caption{\eval Completeness evaluation example for each step based on GPT-4o.}
\label{case:eval_comp}
\end{table*}
\begin{table*}[htbp]
\centering
\begin{small}
\begin{tabularx}{\textwidth}{%
>{\arraybackslash}p{0.15\textwidth}
>{\arraybackslash}X
}
\toprule
\multicolumn{2}{l}{\textbf{\eval Faithfulness : Claude 3.5 Sonnet (score:} 60.53\textbf{) v.s. Llama 3.1-8B (score:} 24.32\textbf{)}} \\ \midrule
\textbf{Tables}
& \textcolor{cyan}{\textbf{Table 1}} \\
& [TITLE] 1970 John Player League - Batting averages - Batting averages \\
& [HEADER] Player | County | Matches | Innings | Runs | Average | Highest\_Score | \_100s | \_50s \\
& [ROW 1] Barry Richards Category:Articles with hCards | Hampshire | 12 | 11 | 592 | 65.77 | 155* | 3 | 0 \\
& [ROW 2] Clive Lloyd Category:Articles with hCards | Lancashire | 14 | 14 | 521 | 57.88 | 134* | 1 | 2 \\
& [ROW 3] Harry Pilling Category:Articles with hCards | Lancashire | 16 | 15 | 625 | 52.08 | 85 | 0 | 5 \\
& ... \\
& \textcolor{purple}{\textbf{Table 2}} \\
& [TITLE] Somerset County Cricket Club in 2010 - Friends Provident t20 - Batting averages \\
& [HEADER] Player | Matches | Innings | Runs | Average | Strike\_rate | Highest\_Score | \_50s \\
& [ROW 1] James Hildreth Category:Articles with hCards | 19 | 19 | 459 | 32.78 | 110.60 | 77* | 2 \\
& [ROW 2] Kieron Pollard Category:Articles with hCards | 17 | 16 | 354 | 32.18 | 175.24 | 89* | 2 \\
& [ROW 3] Marcus Trescothick Category:Articles with hCards | 19 | 19 | 572 | 31.77 | 157.14 | 83 | 6 \\
& ... \\
& \textcolor{purple}{\textbf{Table 10}} \\
& [TITLE] Somerset County Cricket Club in 2009 - Match logs and statistics - Bowling averages \\
& [HEADER] Player | Matches | Overs | Wickets | Average | Economy | BBI | \_4wi \\
& [ROW 1] Ben Phillips Category:Articles with hCards | 8 | 47.0 | 14 | 14.00 | 4.17 | 3/24 | 0 \\
& [ROW 2] Charl Willoughby Category:Articles with hCards | 6 | 42.4 | 11 | 19.18 | 4.94 | 3/36 | 0 \\
& [ROW 3] Zander de Bruyn Category:Articles with hCards | 8 | 32.4 | 6 | 28.33 | 5.20 | 4/20 | 1 \\
& ... \\ \midrule
\textbf{Question}
& What can be inferred about the batting styles and strategies of players in the 1970 John Player League versus those in the Somerset County Cricket Club in 1891 based on their average scores and highest scores? \\ \midrule
\textbf{Claude 3.5 Sonnet}
& \textcolor{cyan}{The data provided does not include information about the Somerset County Cricket Club in 1891, so a direct comparison between that year and the 1970 John Player League is not possible}. However, we can draw insights from the 1970 John Player League data. \textcolor{cyan}{In the 1970 season, batsmen like Barry Richards and Clive Lloyd demonstrated exceptional batting prowess, with Richards averaging an impressive 65.77 runs and Lloyd 57.88 runs}. These high averages, coupled with their ability to score centuries (Richards had 3 and Lloyd had 1), suggest that the top batsmen of that era were capable of both consistency and explosive performances. \textcolor{cyan}{The presence of multiple players with averages above 40 and the ability to score centuries or half-centuries} indicates that the 1970 John Player League featured aggressive batting strategies, with players aiming for high individual scores while maintaining solid averages throughout the season. This style of play likely contributed to entertaining matches and high team totals, reflecting the evolving nature of limited-overs cricket in that period. \\ \midrule
\textbf{Llama 3.1-8B}
& The batting styles and strategies of players in the 1970 John Player League and \textcolor{purple}{the Somerset County Cricket Club in 2010 are distinct from those in the 1969 Player's County League and the Somerset County Cricket Club in 2009, as evidenced by their average scores and highest scores}. In the 1970 John Player League, players like Barry Richards and Clive Lloyd achieved high average scores and highest scores, indicating a more aggressive batting approach. In contrast, \textcolor{purple}{players in the Somerset County Cricket Club in 2010, such as James Hildreth and Marcus Trescothick, had lower average scores and highest scores, suggesting a more defensive approach}. This difference in batting style is also reflected in the \textcolor{purple}{1969 Player's County League, where players like Keith Boyce and Peter Sainsbury had high average scores and highest scores}, indicating an aggressive approach. However, \textcolor{purple}{the Somerset County Cricket Club in 2009, with players like Zander de Bruyn and Craig Kieswetter, had lower average scores and highest scores}, suggesting a more defensive approach. This contrast highlights the evolution of batting styles and strategies over time, with teams and players adapting to different conditions and opponents. \textcolor{purple}{The Somerset County Cricket Club in 2010, with its more defensive \textit{...Repeat similar contents...}} \\ \midrule
\textbf{Comparative}
& \textcolor{cyan}{Claude 3.5 Sonnet’s response clearly establishes the premise that “a conclusion cannot be drawn due} \\
\textbf{analysis}
& \textcolor{cyan}{to the lack of data from 1891”} and maintains consistency by focusing solely on the statistics within the 1970 John Player League. In contrast, \textcolor{purple}{Llama 3.1-8B’s response expands the scope excessively by referencing multiple seasons and tournaments (1969, 2009, 2010, etc.), including information that is not directly relevant to the original question}, which compares 1891 to 1970. As a result, while neither response reaches a perfect conclusion (due to retrieved tables), Claude 3.5 Sonnet’s answer aligns more closely with the intent of the question by acknowledging the limitations of the available data and maintaining a coherent interpretation. \\
\bottomrule
\end{tabularx}
\end{small}
\caption{Comparison of \eval Faithfulness results between Claude 3.5 Sonnet and Llama 3.1-8B.}
\label{case:comparison_i}
\end{table*}
\begin{table*}[htbp]
\centering
\begin{small}
\begin{tabularx}{\textwidth}{%
>{\arraybackslash}p{0.15\textwidth}
>{\arraybackslash}X
}
\toprule
\multicolumn{2}{l}{\textbf{\eval Completeness : GPT-4o (score:} 68.97\textbf{) v.s. Qwen2-7B (score:} 9.84\textbf{)}} \\ \midrule
\textbf{Question}
& Summarize the performance of all Belarusian clubs in the European Cups across the 2006, 2007, and 2008 seasons, focusing on the rounds they reached and the teams they faced in each leg of the competition. \\ \midrule
\textbf{Insight}
& \textcolor{cyan}{Across the 2006, 2007, and 2008 seasons, Belarusian clubs had varied performances in European Cups}. \textcolor{cyan}{In 2006, MTZ-RIPO Minsk participated} in the UEFA Intertoto Cup, losing to Shakhter Karagandy in the first round and subsequently to Moscow in the second round. \textcolor{cyan}{Dinamo Minsk and BATE Borisov also competed} in the UEFA Cup, with Dinamo Minsk eliminated by Artmedia Bratislava and BATE Borisov losing to Rubin Kazan. \textcolor{cyan}{Shakhtyor Soligorsk faced} defeat in the UEFA Champions League against Široki Brijeg. \textcolor{cyan}{In 2007, Shakhtyor Soligorsk advanced} past Ararat Yerevan in the Intertoto Cup but was eliminated by Chernomorets Odessa in the second round. \textcolor{cyan}{Dinamo Brest and Dinamo Minsk had mixed results} in the UEFA Cup, with Dinamo Brest advancing past Liepājas Metalurgs but Dinamo Minsk failing against Odense. \textcolor{cyan}{BATE Borisov faced} Villarreal in the UEFA Cup and was eliminated, while in the Champions League, they had a narrow defeat against APOEL Nicosia but managed to progress past Hafnarfjarðar before losing to Steaua Bucureşti. \textcolor{cyan}{In 2008, Shakhtyor Soligorsk} was eliminated in the Intertoto Cup by Cracovia, while Gomel and MTZ-RIPO Minsk both lost in the UEFA Cup qualifying rounds. \textcolor{cyan}{BATE Borisov had a notable run} in the Champions League, defeating Valur and reaching the group stage, where they faced tough competition from Real Madrid, Juventus, and Zenit St. Petersburg, ultimately finishing with mixed results. \textcolor{cyan}{Overall, Belarusian clubs struggled to advance past the early rounds in most competitions during these seasons, with BATE Borisov achieving the most significant success by reaching the group stage of the Champions League in 2008.} \\ \midrule
\textbf{GPT-4o}
& \textcolor{cyan}{Across the 2006, 2007, and 2008 seasons, Belarusian clubs had mixed performances in European Cups}, with BATE Borisov emerging as the most successful team. \textcolor{cyan}{In 2006, BATE} reached the second qualifying round of the UEFA Cup, defeating Nistru Otaci (3-0 aggregate) but losing to Rubin Kazan (5-0 aggregate). \textcolor{cyan}{Shakhtyor Soligorsk} exited in the first qualifying round of the UEFA Champions League against Široki Brijeg (0-2 aggregate), while \textcolor{cyan}{MTZ-RIPO Minsk advanced} past Shakhtyor Karagandy in the Intertoto Cup first round (6-4 aggregate) but fell to Moscow in the second round (3-0 aggregate). \textcolor{cyan}{In 2007, BATE reached} the third qualifying round of the UEFA Champions League, defeating APOEL Nicosia (3-2 aggregate) and Hafnarfjarðar (4-2 aggregate) before losing to Steaua Bucureşti (4-2 aggregate). In the UEFA Cup, \textcolor{cyan}{Dinamo Minsk} defeated Skonto Riga (3-1 aggregate) in the first qualifying round but lost to Odense (5-1 aggregate) in the second, while Dinamo Brest was eliminated by Liepājas Metalurgs (3-2 aggregate) in the first qualifying round. \textcolor{cyan}{Shakhtyor Soligorsk advanced} to the second round of the Intertoto Cup, beating Ararat Yerevan (4-3 aggregate) but losing to Chernomorets Odessa (6-2 aggregate). \textcolor{cyan}{In 2008, BATE had a historic run} in the UEFA Champions League, progressing from the first qualifying round by defeating Valur (3-0 aggregate), then overcoming Anderlecht (4-3 aggregate) and Levski Sofia (2-1 aggregate) to reach the group stage, where they faced Real Madrid, Juventus, and Zenit St. Petersburg, earning two draws but failing to advance. In the UEFA Cup, \textcolor{cyan}{MTZ-RIPO Minsk} lost to Žilina (3-2 aggregate) in the first qualifying round, while Gomel was eliminated by Legia Warszawa (4-1 aggregate). \textcolor{cyan}{Shakhtyor Soligorsk participated} in the Intertoto Cup, defeating Cracovia (5-1 aggregate) in the first round but losing to Sturm Graz (2-0 aggregate) in the second. \textcolor{cyan}{Overall, BATE's group stage appearance in 2008 marked a significant achievement for Belarusian football, while other clubs struggled to progress beyond the early rounds.} \\ \midrule
\textbf{Qwen2-7B}
& \textcolor{cyan}{In the European Cups across the 2006, 2007, and 2008 seasons, Belarusian clubs displayed a commendable performance, advancing to various stages of the tournaments}. \textcolor{cyan}{In 2006, BATE Borisov} reached the UEFA Champions League group stage, defeating Anderlecht and Juventus, while Shakhtyor Soligorsk was eliminated in the UEFA Intertoto Cup's second round. \textcolor{cyan}{In 2007, BATE Borisov made it} to the UEFA Champions League group stage, overcoming APOEL Nicosia, Hafnarfjarðar, and Steaua Bucureşti, and Shakhtyor Soligorsk advanced to the UEFA Intertoto Cup's second round. \textcolor{cyan}{In 2008, BATE Borisov reached} the UEFA Champions League group stage, facing Real Madrid, Juventus, and Zenit St. Petersburg, and Shakhtyor Soligorsk was eliminated in the UEFA Intertoto Cup's first round. \textcolor{cyan}{These results highlight the competitive spirit and skill of Belarusian clubs in European football}. \\ \midrule
\textbf{Comparative}
& \textcolor{cyan}{GPT-4o provides a thorough breakdown of how Belarusian clubs performed in European competitions} \\
\textbf{analysis}
& \textcolor{cyan}{during the 2006, 2007, and 2008 seasons}, detailing the rounds they reached, the teams they faced, and the exact match results. In contrast, \textcolor{purple}{Qwen2-7B gives a much more general summary, primarily focusing on BATE Borisov while omitting other clubs like Shakhtyor Soligorsk and Dinamo Minsk, leaving out key details} such as specific opponents, progression through different stages, and match outcomes. GPT-4o delivers a far richer response by fully covering each season’s results and matchups, making it a much stronger answer. On the other hand, Qwen2-7B’s response lacks depth due to its narrow focus on BATE and contains inaccuracies (e.g., claiming BATE defeated Juventus in 2006), resulting in a lower score. \\
\bottomrule
\end{tabularx}
\end{small}
\caption{Comparison of \eval Completeness results between GPT-4o and Qwen2-7B.}
\label{case:comparison_ii}
\end{table*}

\end{document}